\documentclass[review]{elsarticle}

\usepackage{lineno}
\usepackage{hyperref}
\usepackage{amsmath,amssymb,amsfonts}
\usepackage{graphicx}
\usepackage{textcomp}
\usepackage{xcolor}
\usepackage{algorithm}
\usepackage{algpseudocode}
\usepackage{svg}
\usepackage{caption}
\usepackage{subcaption}
\usepackage{comment}

\journal{Journal of \LaTeX\ Templates}

\begin{document}

\begin{frontmatter}

\title{Longest Common \textit{Substring} in Longest Common \textit{Subsequence}'s Solution Service: A Novel Hyper-Heuristic}

\author[mymainaddress]{Alireza Abdi\corref{mycorrespondingauthor}}
\cortext[mycorrespondingauthor]{Corresponding author}
\ead{alirezaabdi@iasbs.ac.ir}

\author[mymainaddress]{Masih Hajsaeedi}
\ead{masihsadegh@iasbs.ac.ir}

\author[mymainaddress]{Mohsen Hooshmand}
\ead{mohsen.hooshmand@iasbs.ac.ir}

\address[mymainaddress]{Department of Computer Science and Information Technology,\\ Institute for Advanced Studies in Basic Sciences (IASBS), Zanjan, Iran}

\begin{abstract}
The Longest Common Subsequence \textcolor{black}{(LCS)} is the problem of finding a subsequence among a set of strings \textcolor{black}{that} has two properties of being common to all and is the longest. The LCS has applications in computational biology and text editing, among many others. Due to the NP-hardness of the general longest common subsequence, numerous heuristic algorithms and solvers have been proposed to give the best possible solution for different sets of strings. None of them has the best performance for all types of sets. In addition, there is no method to specify the type of a given set of strings. Besides that, the available hyper-heuristic is not efficient and fast enough to solve this problem in real-world applications. This paper proposes a novel hyper-heuristic to solve the longest common subsequence problem using a \textcolor{black}{novel} criterion to classify a set of strings based on their similarity. To do this, we offer a general stochastic framework to identify the type of a \textcolor{black}{given} set of strings. Following that, we introduce \textcolor{black}{the set similarity dichotomizer ($S^2D$)} algorithm based on the framework that divides the type of sets into two. \textcolor{black}{This algorithm is introduced for the first time in this paper and opens a new way to go beyond the current LCS solvers.} Then, we present a \textcolor{black}{novel} hyper-heuristic that exploits the \textcolor{black}{$S^2D$ and} one of the internal properties of the set to choose the best matching heuristic among a set of heuristics. We compare the results on benchmark datasets with the best heuristics and hyper-heuristics. The results show a higher performance of \textcolor{black}{our} proposed hyper-heuristic in both quality of solutions and run time factors. \textcolor{black}{All supplementary files, including the source codes and datasets, are publicly available on GitHub.\footnote[1]{\href{https://github.com/BioinformaticsIASBS/LCS-DSclassification}{https://github.com/BioinformaticsIASBS/LCS-DSclassification}}}

\end{abstract}

\begin{keyword}
Longest common subsequence\sep Longest common substring\sep Hyper-heuristic\sep Upper bound\sep LCS
\end{keyword}

\end{frontmatter}

\section{Introduction}\label{intro}
The Longest Common Subsequence (\textit{LCS}) problem is a classical computer science problem whose goal is to find the longest possible common subsequence among a set of strings $S$ = $\{\textbf{s}_{1},\textbf{s}_{2},\cdots,\textbf{s}_{n}\}$ over a finite set of alphabet $\Sigma$. Where $n$ is the number of strings, and $|\Sigma|$ = $m$. The term subsequence in the \textit{LCS} means the importance of order, and the consecutive position does not matter. The \textit{LCS} has many applications in computational biology~\cite{smith81,Jiang02,shikder19}, pattern recognition~\cite{lu78}, graph learning~\cite{huang21}, text editing~\cite{kruskal83}, web user clustering~\cite{banerjee01}, trajectories, and similarity matching~\cite{dong18, Ding19}. 
The \textit{LCS} is an NP-hard~\cite{maier78} problem for $n > 2$  and does not have an algorithm to find the exact solution \textcolor{black}{in a reasonable time}. Thus, many researchers have tried to solve the \textit{LCS} with heuristics methods. Plenty of heuristic functions work well on some datasets but not all. For example, \textit{BS-Ex}~\cite{djukanovic19} and $k_{analytic}$~\cite{Abdi2022} heuristic algorithms work well on $ES$ and $ACO$-$Random$ datasets, respectively. 
\textcolor{black}{The above observations indicate that one of the key parameters to solving the LCS problem efficiently is the type of datasets.} In another work~\cite{Abdi2022}, we have shown the impact of understanding the \textcolor{black}{type of sets} to have the best possible configuration of our heuristic function. 
\textcolor{black}{In addition, specifying the type of a given set of strings avoids running all heuristics for the set to obtain the best possible solution. Nevertheless, there is no automatic way to obtain the type of a set.}
In the \textit{LCS} literature, researchers have divided types of strings into two classes, correlated and uncorrelated. \textcolor{black}{However,} reporting the type of datasets has been due to being aware of their generation~\cite{blum07}.
\textcolor{black}{In other words}, without knowing the dataset generation, there has been no \textcolor{black}{algorithm} to identify the type of a given set of strings. In addition, in real-world problems, we do not know the type of strings in advance, and obviously, it is not efficient to test all methods.
To have higher performance, it is a must to classify \textcolor{black}{sets}, and it is also necessary to pick the best matching algorithm for each \textcolor{black}{type}. In this work, as one of our contributions, we propose a framework and an algorithm to identify the type \textcolor{black}{of sets for the first time}. This identification helps \textcolor{black}{us to design a novel hyper-heuristic} to find the best heuristic among a set of heuristics.

The contribution of this paper is three-fold. As the first contribution, the paper proposes a general stochastic framework to examine the similarity among the strings of a given set. Regarding the framework, \textcolor{black}{we introduce our second contribution ($S^2D$ algorithm), a fast, simple algorithm that classifies any given set of strings and identifies its type. This algorithm is the first classification algorithm of sets of strings in the LCS literature.} The last contribution of this paper is a novel hyper-heuristic based on the \textcolor{black}{$S^2D$} algorithm \textcolor{black}{and one of the internal properties of strings} for the \textit{LCS} problem. \textcolor{black}{The proposed hyper-heuristic performs well and is time-efficient compared to the previously proposed hyper-heuristic and best methods for the LCS problem.}

The rest of this paper comes as follows. Section~\ref{req} defines basic notations and preliminaries. Section~\ref{relw} addresses the related work. We have divided this section into three subsections. \textcolor{black}{Subsection~\ref{sec:sos}} deals with the concepts of similarity among strings. Subsection~\ref{sec:relw-hh} describes the only available hyper-heuristic for the \textit{LCS} problem and its properties. \textcolor{black}{The last subsection (\ref{sec:base}) describes} \textcolor{black}{base heuristic functions for our hyper-heuristic, and the upper bound function}. \textcolor{black}{In section~\ref{proposal}, we introduce our contributions in three subsections. Subsection~\ref{proposal:fw} provides the string classifier framework. After that, subsection~\ref{sec:clss_alg} presents the $S^2D$ algorithm that classifies sets of strings into two types, correlated and uncorrelated. Using the mentioned classifier, we propose a novel hyper-heuristic to solve the \textit{LCS} problem in subsection~\ref{sec:hh}.} Section~\ref{res} shows the experimental results and comparison of the proposed methods with the best methods \textcolor{black}{ from different aspects. Section~\ref{sec:statsign} provides a statistical analysis of the results.} Section~\ref{conc} finally concludes the paper.

\section{Basic definitions and preliminaries}\label{req}
In this section, we define the basic definitions of our work and some useful functions for a better understating of the following sections. Before all else, Table~\ref{tab:abbr} presents all the symbols used in the paper.
\begin{table}[]
    \centering
     \caption{List of symbols}
    \begin{tabular}{lll}
\hline
    \textbf{Symbol} & \textbf{Meaning} & \textbf{Type}\\ \hline
     S & set of strings  & set \\
     \textbf{s} & a string, $s\in S$&string \\
     $\Sigma$ & alphabet set&set \\
     $\sigma$ & a member of alphabet set, $\sigma\in\Sigma$ & character\\
     n & number of strings& scalar\\
     m & number of alphabets &scalar\\
     $\ell$ & length of string \textbf{s} &scalar\\
     $\textit{LCS}$ & longest common subsequence& string \\
     $\textit{LCT}$ & longest common substring& string \\
     $\beta$ & beam width &scalar\\
     $\beta_{h}$ & trial beam width& scalar\\
     $L_C$ & list of children potential for expansion in beam search&set\\
     $L_B$ & list of best children chosen for next level in beam search&set\\
     $\text{o}_{i}^{(\sigma)}$ & number of occurrence of character $\sigma$ in s$_{i}$ & scalar\\
     \textbf{r}$_i^{\sigma}$ & postfix of s$_{i}$ after deleting character $\sigma$ &string\\
     R$^{\sigma}$ & list of all \textbf{r}$_i^{\sigma}$ &set\\
     $\nu$ & a node in beam search& node\\
     S$_{sub}$ & a subset of S &set\\
     \textit{f}$_{sim}$ & similarity evaluation function &function\\
     S$_{sub}^{partition}$ & random substrings set from S$_{sub}$& set\\
     sim$_{sub}$ & the similarity value of S$_{sub}^{partition}$&scalar\\
     Sim$_{S}$ & list of all sim$_{sub}$s &set\\
     ei & end index&scalar\\
     si & start index& scalar\\ \hline
\end{tabular}
    \label{tab:abbr}
\end{table}

$S = \{ \textbf{s}_1, \textbf{s}_2, \cdots, \textbf{s}_{n} \}$ is the set of strings over an alphabet set $\Sigma = \{ \sigma_1, \sigma_2, \cdots, \sigma_m \}$, where $n$ is the number of strings and $m$ is $|\Sigma|$. In this paper, we use beam search~\cite{norvig92} as our search strategy. We choose beam search due to its high performance in solving the \textit{LCS} problem~\cite{blum09}.
Beam search is a limited version of breadth first search, which at each level expands $\beta$ nodes instead of all nodes. Its way of working is to start with the root node and put it in the \textit{list of best children} (\textit{$L_B$}). 
In each step, the algorithm expands the children nodes of the available nodes in the \textit{$L_B$} and puts them in a list called the \textit{children node list} (\textit{$L_C$}). Then, the algorithm applies a score or heuristic function on each node in \textit{$L_C$} and computes their value to locate the best possible paths to \textit{LCS}.
Finally, it chooses the $\beta$ most promising nodes among \textit{$L_C$} and puts them in \textit{$L_B$} for next-level expansion. This procedure iterates until the \textit{$L_B$} list becomes empty.

Beam search specifies each node by $\nu$, equivalent to choosing an alphabet. Each node $\nu$ has its own set of $R^\nu$. The set $R^\nu$ is the remainder of all strings after the deletion alphabet $\nu$ and its prefixes, $|R^\nu|=n$. The symbol $o_{i}^{(\sigma_{j})}$ is the cardinality of the unique alphabet $\sigma_j$ in the \textit{i-th} string, where $1 \leq i \leq n$ and $1 \leq j \leq m$. For example, suppose $S$ is $\{ \texttt{TGACTGCA}, \texttt{GACTTGAG} \}$, which its $o_{1}^{(\texttt{A})}=2$. Here, $R^{\texttt{C}}$ is $\{$ \textbf{r}$_1 = \texttt{TGCA}, $\textbf{r}$_2 = \texttt{TTGAG} \}$.

As mentioned above, beam search chooses the $\beta$ best nodes in each level from $L_C$. This sieving takes advantage of a heuristic function. Most promising heuristics assign the function $p(k,\ell)$, representing the probability of having a random subsequence with length $k$ from $s$ with length $\ell$~\cite{mousavi12improved, Abdi2022}.

In this paper, we take advantage of the Longest Common subsTring (\textit{LCT}) to \textcolor{black}{identify the type of a given set}. So, it is necessary to describe it briefly. The \textit{LCT} is the problem of finding the longest substring among a set of strings $S$. It is worth mentioning the difference between substring and subsequence. A substring of string $s$ is acquired consecutively. A subsequence of string $s$ is acquired successively~\cite{dragon}. For example,  for set $S$ = $\{\texttt{heaaebdgbc}, \texttt{heaabdbcde},$ $\texttt{heaaebdgbh}\}$,   the string $\texttt{heaa}$ is an \textit{LCT}, and the string $\texttt{heaabdb}$ is an \textit{LCS}.

\section{Related work}\label{relw}
In this section, we first describe those works that addressed the type of \textcolor{black}{sets in subsection~\ref{sec:sos}}. Reading those works makes clear that despite mentioning the string types, there is no algorithm to find the type of strings. Then, we explain the only hyper-heuristic proposed for the \textit{LCS} problem \textcolor{black}{in section~\ref{sec:relw-hh}}. Lastly, we present \textcolor{black}{the upper bound function as an internal property for a set and the base heuristic functions for our proposed hyper-heuristic}.

\subsection{Type of sets}\label{sec:sos}
\textcolor{black}{Mostly, the researchers divided sets into two types and used different criteria (\textit{i.e.}, correlated vs. uncorrelated or uniform vs. non-uniform) to dichotomize them.}
All datasets had been uncorrelated in the history of the \textit{LCS} problem. For the first time, Blum and Blesa~\cite{blum07} produced the ``\textit{BB}'' dataset as a correlated dataset. They have divided the types of \textcolor{black}{sets} into correlated and uncorrelated types. In 2021, Nikolic et al.~\cite{Nikolic21} divided \textcolor{black}{sets} into two \textcolor{black}{types:} ``uniform'' and ``nonuniform''. 
The dichotomy between these two types roots in the frequency of unique alphabets --- the \textit{uniform} \textcolor{black}{sets} are those with uniform distribution of the number of unique alphabets, and the \textit{nonuniform} \textcolor{black}{sets} are those without uniform distribution. Their definition for distinguishing the \textit{LCS} benchmarks misses some facts and is not a complete and adequate criterion in \textcolor{black}{some} cases. As the authors mentioned in their paper, uniformity cannot determine the category of the ``\textit{BB}'' dataset.
Indeed, the cardinality distribution of characters in \textit{BB} is uniform, but the \textit{LCS} of this benchmark cannot be suitably located by methods designed for uniform strings. In addition to the author's description about \textit{BB}, the \textit{ACO-Rat}, reversely, if we borrow the term \textit{uniform} from~\cite{Nikolic21}, is a nonuniformly distributed dataset. Still, it is solvable by those methods \textcolor{black}{with} high performance for uniform datasets. These facts demonstrate that uniform-nonuniform classification misses the point in \textcolor{black}{some} cases.
Thus, in this paper, we return to the correlated-uncorrelated dichotomy \textcolor{black}{among} sets of strings. \textcolor{black}{However, there is no algorithm to automatically classify a set of strings into two correlated or uncorrelated types, and identifying the type of sets has roots in knowing the way of their generation.}

\subsection{Hyper-heuristics for the LCS}\label{sec:relw-hh}
Hyper-heuristics are broadly concerned with efficiently choosing the best heuristic for a problem~\cite{burke2003}. For the \textit{LCS} problem, Mousavi and Tabataba~\cite{tabataba12hyper} introduced the only hyper-heuristic that each set of strings is tested with all heuristic functions with small beam width ($\beta_h$), trial beam width. The winner --- the heuristic with the best result among all heuristics --- is chosen as the scoring function but with a bigger beam width or, equivalently, the actual beam width ($\beta$). The difference between $\beta_{h}$ and $\beta$ is in their size; $\beta_{h}$ is always smaller than $\beta$. We call this hyper-heuristic ``trial $\&$ error'' hyper-heuristic (TE-HH).
Aside from its advantages, TE-HH has some drawbacks. First, each heuristic function is separately tested with $\beta_h$, causing redundancy and computational overhead. Secondly, choosing proper $\beta_{h}$ is a challenging task. Although when a heuristic obtains a better result for the smaller $\beta$, it will not guarantee that the bigger $\beta$ has a higher result. In other words, the size of the the $\beta$ is not a neutral variable. It can be called the neutrality fallacy. 

\subsection{Base heuristic functions}\label{sec:base}
\textcolor{black}{A hyper-heuristic has some base heuristic functions. Here, we describe the base heuristics for our proposed hyper-heuristic, which we introduce in section~\ref{sec:hh}.}

\textcolor{black}{The first base} heuristic is $\textit{BS-Ex}$, introduced by Djukanovic et al.~\cite{djukanovic19} \textcolor{black}{to solve} the \textit{LCS} problem. It achieves high performance for the $ES$ dataset. Its equation for node evaluation is as follows. 
\begin{equation}
    \textit{BS-Ex}(R^\nu) = \sum_{k=1}^{\ell_{min}} (1 - ( 1 - \prod_{i=1}^{m} p (k, |\text{r}_i|)^{\vert \Sigma \vert^{k}}),
    \label{eq:bsex}
\end{equation}
where $\ell_{min}$ is the minimum length of strings in S. \textcolor{black}{The second base} heuristic \textcolor{black}{of our hyper-heuristic} is $k_{analytic}$, which we introduced in~\cite{Abdi2022}. The $k_{analytic}$ has two versions. \textcolor{black}{One is for correlated ($k_{analytic}^{cor}$) and one for uncorrelated ($k_{analytic}^{uncor}$) sets}. The following \textcolor{black}{equations show} the score calculation for a node $\nu$ \textcolor{black}{and how to set parameter $k$}.

\begin{equation}
    h(R^\nu) = \prod_{i = 1}^{n} p(k, |\text{r}_i|).
    \label{eq:score}
\end{equation}
The parameter $k$ in Eq.~\ref{eq:score} is set by Eq.~\ref{eq:uniformeq} for uncorrelated strings and Eq.~\ref{eq:nonuniformeq} for correlated strings.
\begin{equation}
    k_{analytic}^{uncor} = \frac{max\{|R_i^{\nu}|\} * (a - \left( b \times {\ln n} \right) )}{|\Sigma|},
    \label{eq:uniformeq}
\end{equation}
\begin{equation}
    k_{analytic}^{cor} = \frac{min\{|R_i^{\nu}|\} - c}{|\Sigma|},
    \label{eq:nonuniformeq}
\end{equation}
where, $a = 1.8233, b = 0.1588$, $c = 31$, and $i=\{1,...,|L_C|\}, \nu \in L_C$. The last \textcolor{black}{base} heuristic is $GCoV$ which is based on the \textit{coefficient of variation}~\cite{cov98}. We have introduced this heuristic in~\cite{Abdi2022}. The equation of $GCoV$ is as follows. 

\begin{equation}
    GCoV(R^\nu) = \frac{ \left[\mu(R^{\nu})\right]^{2}}{\left[var(R^{\nu})\right]^{\gamma}} \times \sqrt{ub(R^{\nu})},
    \label{eq:cov}
\end{equation}
where, $\mu(\cdot)$ and $var(\cdot)$ are the mean function and the variance function, respectively, the parameter $\gamma$ depends on the number of strings and is  equal to $(0.0036 \times n) - 0.0161$. \textcolor{black}{These three heuristics, i.e., $\textit{BS-Ex}$, $k_{analytic}$, and $GCoV$, are the base heuristic functions in our proposed hyper-heuristic.}

\textcolor{black}{In addition to the above heuristics, there is an upper bound (\textit{ub}) function, which is introduced by Blum et al.~\cite{blum09} to evaluate partial solutions. We use the \textit{ub} function as a measure in our proposed hyper-heuristic in section~\ref{sec:hh}.} It finds the maximum possible length of the \textit{LCS} for $S$ as follows.
\begin{equation}
    ub(R^\nu) = \sum_{j=1}^{m} min \{ o_{i}^{(\sigma_{j})} \mid i = 1, \cdots, n \},
    \label{eq:ub}
\end{equation}
\section{Methods}\label{proposal}
Each of the base heuristics we have introduced in section~\ref{sec:base}, i.e., \textit{BS-Ex}, $k_{analytic}$, and \textit{GCoV}, works well on different types of \textcolor{black}{sets}. For example, \textit{BS-Ex}, as mentioned above, works well for the $ES$ dataset. \textcolor{black}{The main reason behind the above observations is the type of datasets.}
\textcolor{black}{Thus, automizing the type identification of a given set is the first stage to solving the LCS problem efficiently.}
To do this, in section~\ref{proposal:fw}, we propose a general stochastic framework to classify a given set of strings. \textcolor{black}{As a practical case of the proposed framework,} in section~\ref{sec:clss_alg}, we propose an algorithm to classify a set of strings into two correlated and uncorrelated \textcolor{black}{classes for the first time. After identifying the type of a given set, choosing the best matching heuristic is the next stage to solve the LCS problem efficiently. Section~\ref{sec:hh} combines these two stages by proposing a novel hyper-heuristic (\textit{UB-HH}) based on the upper bound function.}

\subsection{Set classification framework}\label{proposal:fw}
\textcolor{black}{First, for the $LCS$ problem, classifying a given set of strings implies evaluating the similarity among its strings to label it appropriately.} These labels help to choose the best matching algorithms for the dataset in real-world problems. In other words, the label is a measure or criteria on strings that declares their total amount of similarities to each other. This similarity can be derived from many properties, from the strings' length and characters' distribution to more complicated measures. The similarity measure can classify sets into two or more \textcolor{black}{types}.
Notably, this framework is not a hyper-heuristic because it does not choose a heuristic to use but classifies strings. In any event, we can use it in hyper-heuristics to select the best possible heuristics across a set of $LCS$ problem solvers. \textcolor{black}{Our proposed} similarity measurement framework relies on two pillars. First, when there exists a similarity among strings of a given set, this similarity will appear in all parts of strings---not only in a specific part of strings. Secondly, the resemblance must exist in all strings---not a subset of strings. 

\begin{figure}
  \centering
  \includegraphics[width=0.9\textwidth]{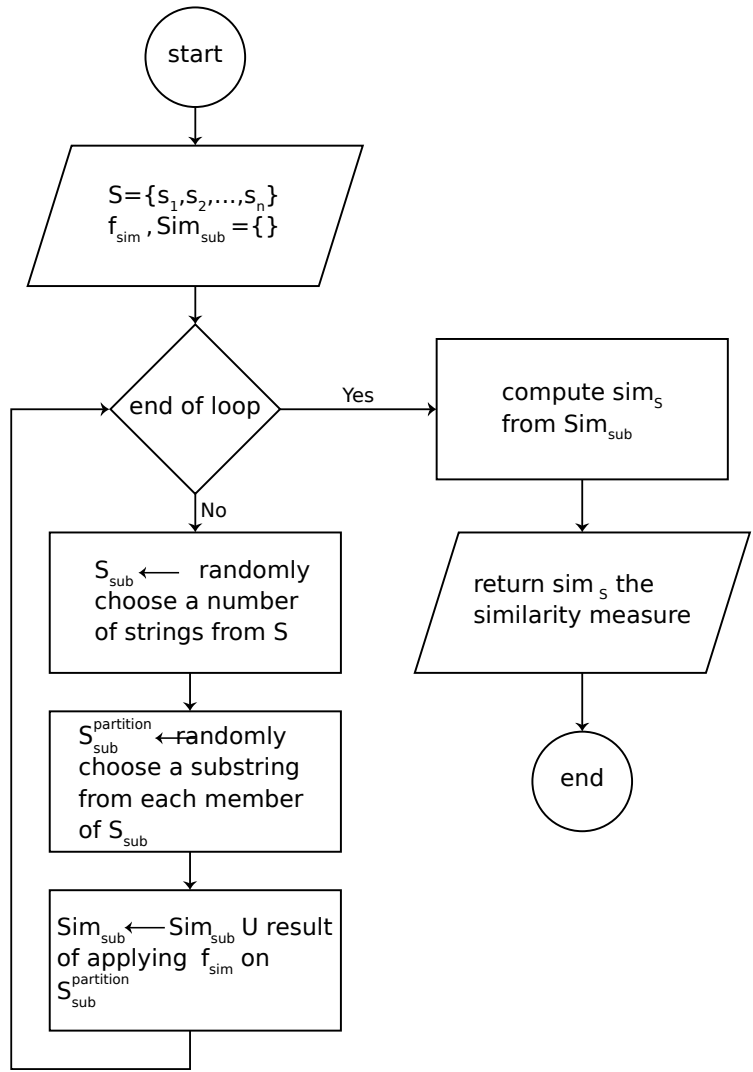}
   \caption{\textcolor{black}{The \textit{SCF} flowchart}. The input is the set of strings, and its output is the amount of similarity among members of $S$. In each iteration, the $SCF$ chooses some substrings from sets and computes their similarity using $f_{sim}$. In the end, the total similarity of $S$ is returned. The similarity value is the measure of classifying $S$.}
  \label{fig:FL}
\end{figure}

Checking all parts of all strings takes too much time, so we decide to design the framework in a reliable stochastic format. We introduce the Stochastic Classification Framework ($SCF$) with the above assumptions. As figure~\ref{fig:FL} presents, the \textcolor{black}{$SCF$} input is a set of strings $S$.
In addition to the mentioned set, defining the similarity evaluation function $f_{sim}$ as another input is necessary. The mentioned function is a measure function that computes the similarities among members of $S$. Utilizing the function $f_{sim}$, the \textcolor{black}{$SCF$} randomly chooses an arbitrary number of strings and puts them in a subset $S_{sub}$ from the set $S$ in each iteration.
In other words, the specifications of $f_{sim}$ dictate the number of strings we pick randomly. Then, the \textcolor{black}{$SCF$} extracts a random partition---substring---from each member of $S_{sub}$ and \textcolor{black}{puts} them in the $S_{sub}^{partition}$ set. Following that, the $S_{sub}^{partition}$ is fed to the $f_{sim}$.
The output of the $f_{sim}$ is the similarity value of strings in $S_{sub}^{partition}$ and it is added to the current set of such values $Sim_{sub}$. The \textcolor{black}{$SCF$} repeats the procedure in several iterations. When the loop ends, the \textcolor{black}{$SCF$} checks all acquired similarity values saved in the $Sim_{sub}$, applies a statistical function, and returns a value label $sim_S$.
The output value is a prediction of the similarity of $S$. In short, the \textcolor{black}{$SCF$} aims to classify the sets based on their similarity values. Thus, as mentioned earlier, we can classify the sets into two or more classes.
\textcolor{black}{This} framework provides an efficient way to check a set's similarity dose. It leads to assigning an \textit{LCS} algorithm to solve the \textit{LCS} problem more efficiently.

\subsection{$S^2D$ algorithm for set classification}\label{sec:clss_alg}
The previous section proposes the $SCF$ to classify a set of strings based on their similarity. This section proposes Set Similarity Dichotomizer ($S^2D$) as an algorithm based on the $SCF$. The $S^2D$ algorithm utilizes \textit{LCT} as $f_{sim}$ to evaluate the similarity of strings. Moreover, it classifies sets into two \textcolor{black}{types} ``correlated'' and ``uncorrelated''. 

Algorithm~\ref{alg:lcsst} shows the steps of the $S^2D$ algorithm. \textcolor{black}{$S$ and the} parameters, \textit{LCT}, and \textit{end-index} ($ei$) are algorithm inputs and represent a set of strings, $f_{sim}$, and the hyper-parameter of the algorithm, respectively. In each iteration, the procedure randomly chooses two strings, $\textbf{s}_i$, and $\textbf{s}_j$, from $S$, where  $1 \leq i, j \leq n$, and $i \neq j$. We define $l_{min}$ as the minimum length of $\textbf{s}_i$ and $\textbf{s}_j$. Then, we separate a part of $\textbf{s}_i$ and $\textbf{s}_j$ using two \textit{start-index} (\textit{si}) and $ei$ parameters. $si \in [1, (l_{min} - ei)]$ is a random number. In fact, $si$ and $ei$ \textcolor{black}{are required} to choose a random substring from $\textbf{s}_i$ and $\textbf{s}_j$. After that, we pass $\textbf{s}_i[si:si + ei]$ and $\textbf{s}_j[si:si + ei]$ as $S_{sub}$ to the $LCT$ function and obtain the length of \textcolor{black}{the longest common} substring, $x$. 
\textcolor{black}{$S^2D$ iterates this process until the convergence. This iteration is necessary to be assured of the $S^2D$ result's reliability. To show the importance of the iteration, we assume the correlated set of  $S = \{s_1 = \texttt{CCGTGCATTT}$, $s_2 = \texttt{CCGTGCGGCA}, s_3 = \texttt{ACGTGCATT}$, $s_4 = \texttt{ACGTGCATT}$, $s_5 = \texttt{CCGTGAATTT}$, $s_6 = \texttt{CCGTGCATAT}$, $\}$. There is a high correlation among the strings of $S$. Thus, it needs to be labeled as correlated. Nevertheless, if we repeat the $S^2D$ in just a single iteration, it is possible to choose the last parts of $s_1$ and $s_2$ (i.e., $\texttt{CATTT}$ and $\texttt{CGGCA}$). Then, it labels $S$ as uncorrelated, which is wrong. To avoid this issue, $S^2D$ repeats the procedure for several iterations and reduces the probability of such problems. According to our extensive tests, repeating the $S^2D$ for $[\frac{n}{2}]$ times guarantees its correctness.}
Finally, the mean of the $Sim_{sub}$ list is compared to a threshold, and consequently, its \textcolor{black}{type}---correlated or uncorrelated---is revealed. 

\begin{algorithm}[H]
\caption{$S^2D$ algorithm for classifying sets.}
\label{alg:lcsst}
\begin{algorithmic}
\Require $S=\{\textbf{s}_1,\textbf{s}_2,\cdots, \textbf{s}_n\}$, $LCT(\cdot,\cdot)$, $ei$
\State $Sim_{sub} = \{\}$
\While {Not Converged}
    \State $i, j = random(1, n)$ \Comment{choose a random number between 1 and $n$}
    \State $l_{min} = min(|\textbf{s}_i|,|\textbf{s}_j|)$ 
    \State $si = random(1, \ell_{min} - ei)$
    \State $x$ = $LCT(\textbf{s}_i[si:si + ei], \textbf{s}_j[si:si + ei])$
    \State $Sim_{sub} = Sim_{sub} \cup \{x\}$
\EndWhile
\State $sim_s$ = $\mu(Sim_{sub})$
\If {$sim_s > tr$}
    \State \Return ``correlated''
\Else
    \State \Return ``uncorrelated''
\EndIf
\end{algorithmic}
\end{algorithm}
\subsection{A novel upper bound hyper-heuristic}\label{sec:hh}
\begin{figure}
  \centering
  \includegraphics[width=0.9\textwidth]{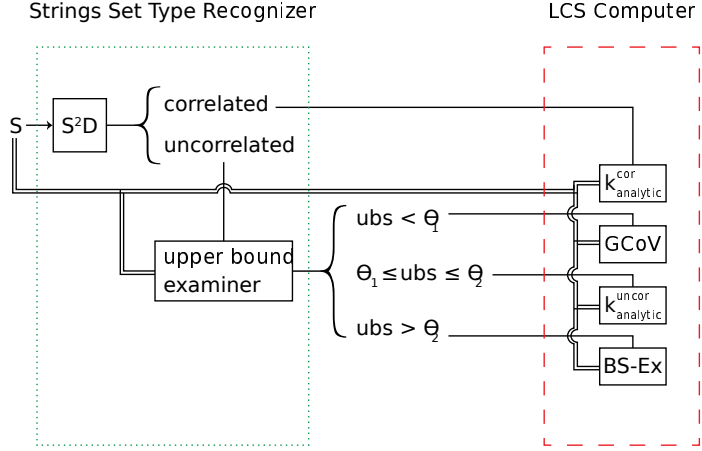}
   \caption{The \textit{UB-HH} algorithm. It consists of two parts: Strings' Set Type Recognizer (green dotted box) and LCS Computer (red dashed box). The former contains two modules—the first module ($S^2D$) examines the set of strings to identify whether they are correlated. If stamped as uncorrelated, the second module computes the upper bound of the set. Based on the information from the first part, one of the heuristic functions in the second part is activated to compute the LCS. }
  \label{fig:UBHH}
\end{figure}

As mentioned earlier, \textcolor{black}{while each heuristic has a high performance on some datasets, it has no proper performance for some other datasets}. Our observations and experiments show that the upper bound size ($ubs$) is an important property \textcolor{black}{in identifying} the goodness of a heuristic function on a set of strings.
By introducing the $S^2D$ algorithm and having $ubs$, we define a novel hyper-heuristic and name it \textit{UB-HH} \textcolor{black}{in this paper}. It has several advantages. First, it avoids the redundancy of the previous hyper-heuristic, \textit{TE-HH}.
\textcolor{black}{Second,} while the \textit{TE-HH} is designed for just beam search, the \textit{UB-HH} can be used in any search strategy (e.g., $A^*$, greedy search) for \textcolor{black}{solving the LCS problem}. \textcolor{black}{Thirdly, it is light and fast in comparison with the \textit{TE-HH}.} Figure~\ref{fig:UBHH} illustrates the \textit{UB-HH} schema.
The \textit{UB-HH} consists of two parts. We refer to the first part as Strings Set Type Recognizer, shown in a green dotted box. It receives the set of strings, and its mission is to determine the set's type. The second part, surrounded by a red dashed box, contains a set of $LCS$ heuristics. For each given set, \textit{UB-HH} activates precisely one of the $LCS$ heuristics.
The activated heuristic computes the $LCS$ of the given set and returns it as the output of the \textit{UB-HH} algorithm. Below, we explain the components of the \textit{UB-HH} in more detail. 
The Strings Set Type Recognizer part consists of two modules. The first module is the $S^2D$ algorithm which determines the class of a given set as correlated and uncorrelated. 
The correlated strings are directly fed to $k_{analytic}^{cor}$ in the second part of \textit{UB-HH} to compute its $LCS$. If the given set receives the uncorrelated label, it will be fed into the upper bound examiner module. 
The upper bound examiner calculates the $ubs$ of the given set using Eq.~\ref{eq:ub} and divides it by $min(|\textbf{s}_i|)$ to map the $ubs$ into the interval $[0, 1]$.
Then, due to having three base heuristic functions for uncorrelated strings, i.e., \textit{BS-Ex}, $k_{analytic}^{uncor}$, and $\textit{GCoV}$, we define two thresholds, $\theta_1$ and $\theta_2$, to divide the output of the upper bound examiner into three intervals. \textcolor{black}{Based on the result of \textit{ubs}, one of them is chosen to find the LCS.}

\subsection{Complexity analysis of UB-HH}
The \textit{UB-HH} first checks the correlation of the $n$ strings using $S^2D$. As algorithm~\ref{alg:lcsst} shows, 
the time complexity of the main loop is O(\textit{n}). The operations in the loop except for $LCT(\cdot,\cdot)$ are O(1). We use dynamic programming to solve the \textit{LCT} problem, and its complexity is the multiplication of the length of two input strings~\cite{gusfield97}.
Here, the length of our strings is $ei$. So, the complexity of the \textit{LCT} procedure is $O(ei^{2})$. Thus, the time complexity of the loop is equal to $O(n\times ei^{2})$. The mean function is out of the loop and its time complexity is O(\textit{n}). Thus, the total complexity of  $S^2D$ is $O(n\times ei^{2}+n)$. 

The \textit{UB-HH} runs the upper bound examiner module for the uncorrelated strings. The upper bound examiner computes the $ubs$ of the strings, which is equal to $O(n\times \ell)$ in the worst case. Thus, the complexity of the \textit{UB-HH} algorithm in the worst case is $O(n\times ei^{2}+n\times \ell + n)$. Because complexity deals with the larger terms, then its whole complexity is either $O(n\times ei^{2})$ or $O(n\times \ell)$---it depends on whether $ei^{2}>\ell$ or vice-versa.

\textcolor{black}{We should note that \textit{TE-HH} needs to run all heuristics with a smaller beam width ($\beta_h$) in advance to find the best LCS heuristic among them. In other words, its time complexity depends on the number and complexity of each base heuristic it uses. We have compared the run time of the \textit{UB-HH} with the \textit{TE-HH} in section~\ref{sec:time}.}

\section{Results}\label{res}
\textcolor{black}{This section provides the results of comparing our proposed methods with the best methods in the LCS literature. To do this, we use three scenarios for comparing methods, \textit{low-time}, \textit{balanced-time-quality}, and \textit{high-quality}. The difference among these scenarios is in their beam widths, which beam width ($\beta$) of \textit{low-time}, \textit{balanced-time-quality}, and \textit{high-quality} are 50, 200, and 600, respectively.
This paper proposes a set classifier, $S^2D$, and a novel hyper-heuristic \textit{UB-HH} method. The $S^2D$ algorithm classifies sets of strings for the LCS problem. To evaluate our proposed methods, we use six benchmark datasets~\cite{shyu09, Easton08, blum07}, $ACO$-$Random$, $ACO$-$Rat$, $ACO$-$Virus$, $SARS$-$CoV$-$2$, $BB$, and $ES$. $BB$ and $SARS$-$CoV$-$2$ datasets belong to the correlated class.
The \textit{ACO-Random}, \textit{ACO-Rat}, \textit{ACO-Virus}, and $ES$ datasets belong to the uncorrelated class. Each of these datasets contains several sets of strings. Each set has different properties based on the alphabet size ($|\Sigma|$), the length of its corresponding strings ($l$), and the number of strings ($n$).
To the best of our knowledge, \textit{BS-Ex}, \textit{BS-GMPSUM}, $k_{analytic}^{uncor}$, $k_{analytic}^{cor}$, and \textit{GCoV} reported the best results on benchmark datasets. So, we compare the \textit{UB-HH} method with the mentioned methods. In addition to compare the \textit{UB-HH} with the best methods, we have replaced the base heuristics of \textit{TE-HH} with the base heuristics used in \textit{UB-HH}. This replacement yields the comparison of both hyper-heuristics in the same condition. To have a fair comparison, we have re-implemented \textit{BS-Ex}, \textit{BS-GMPSUM}, $k_{analytic}^{uncor}$, $k_{analytic}^{cor}$, \textit{GCoV}, and \textit{TE-HH} methods in \textit{python} version $3.8$ programming language.
Thus, this section contains the following subsections. Subsection~\ref{sec:tune} reports the way of tuning the algorithms' parameters in the simulation. Then, we report the performance of the $S^2D$ algorithm in section~\ref{ress2d}. Then, to show the performance of the \textit{UB-HH} algorithm, we compare it with the best heuristics in the LCS literature in section~\ref{resubhh}. subsection~\ref{sec:time} provides a run time comparison between \textit{TE-HH} and \textit{UB-HH}.}

\subsection{Parameter tuning}\label{sec:tune}
\textcolor{black}{
To tune parameters, we have generated several new correlated and uncorrelated datasets with different sizes of $|\Sigma|$ in the way that Blum and Blesa~\cite{blum07} generated the \textit{BB} dataset. In other words, these generated datasets are just for parameter tuning. 
The mutation (deletion, insertion, change) probability for correlated and uncorrelated is set to 0.1 and 0.9, respectively. For more detail about generating datasets, see~\cite{blum07}. Using grid search~\cite{fayed19}, we tested different values of \textit{ei} $= \{10, 15, \cdots, \ell \}$, and \textit{tr} ($1 < tr < ei$) for generated datasets. Our extensive experiments revealed that if a set has a common substring with a length greater than $5$, the set in question is correlated.
If the length of the common substring becomes lower than $5$, the set is uncorrelated. So, we set the threshold for distinguishing datasets type to $5$. Also, we have observed that a value of \textit{ei} $\geq 20$ is suitable for the algorithm to work properly. To reduce the algorithm's run time, we set the value of $ei = 20$. When $|\Sigma| = 2$, the chance for a common substring with a length greater than $5$ for both types (i.e., correlated and uncorrelated) is high, and all sets are identified as correlated.
To address this issue, we reduced the size of \textit{ei} for sets with $|\Sigma| = 2$. To make the long story short, we set \textit{ei} = 20 and $tr$ = 5 for $|\Sigma| > 2$, and \textit{ei} = 10 and $tr$ = 5 for $|\Sigma| = 2$. For tuning the upper bound examiner module parameters (i.e., $\theta_1$ and $\theta_2$), we tested different values of $\theta$ from a list ---$[0.01, 0.02, \cdots, 1]$--- and obtained the best values of $\theta$, which for $\theta_1$ is $0.54$ and for $\theta_2$ is $0.9$.}

\subsection{$S^2D$ for classifying benchmarks}\label{ress2d}
\textcolor{black}{As introduced in section~\ref{sec:clss_alg}, the $S^2D$ algorithm classifies any given set of strings for the LCS problem and identifies its label. To the best of our knowledge, $S^2D$ is the first case of this type of set classification algorithm in the LCS literature. Thus, there is no other algorithm in this area to compare. Therefore, this section provides the result of applying the $S^2D$ algorithm on all mentioned benchmark datasets. Each of the strings' sets is a sample for $S^2D$. This algorithm is applied to all samples provided by datasets, i.e., 751 sets}. The $S^2D$ classifies instances into two sets of correlated and uncorrelated.
Table~\ref{tb:s2d} illustrates the confusion matrix resulting from applying $S^2D$ to the benchmark datasets. As the table shows, there are 91 correlated sets, and $S^2D$ identified all of them correctly. There are 660 uncorrelated sets, of which the $S^2D$ labels 650 correctly and ten wrongly. So, the $S^2D$'s accuracy is $98\%$. The wrong labels belong to those sets with an alphabet length equal to 2. The problem with $|\Sigma|=2$ is that existing of the exact binary substrings has a high probability.

\begin{table}[H]
\centering
\caption{The confusion matrix of applying $S^2D$ on all sets available in $ACO$-$Rat$, $ACO$-$Virus$, $ACO$-$Random$, $ES$, $BB$, and $SARS$-$CoV$-$2$ datasets.}
\begin{tabular}{l|l|l}
\hline
             & Correlated & Uncorrelated \\
             \hline
Correlated   &      91      &      10        \\ \hline
Uncorrelated &      0       &         650    \\
\hline
\end{tabular}
\label{tb:s2d}
\end{table}

\subsection{Comparison of \textit{UB-HH} with best methods}\label{resubhh}
\textcolor{black}{Here, we compare the performance of the \textit{UB-HH} algorithm with the best methods in three mentioned scenarios over six mentioned benchmarks. Subsection~\ref{sec:balance} provides the result of \textit{balanced-time-quality}, or $\beta=200$. In addition to the \textit{balanced-time-quality} scenario ($\beta=200$), we provide two more scenarios of \textit{low-time} ($\beta=50$) and \textit{high-quality}  ($\beta=600$) in the following subsections to fairly compare our proposal with existing methods in the literature. We call the $\beta = 200$ as \textit{balanced-time-quality} due to having both balances in time and quality simultaneously compared to the \textit{low-time} and \textit{high-quality}. Therefore, subsection~\ref{sec:low} reports the \textit{low-time} scenario, or $\beta=50$. Subsection~\ref{sec:high} contains the result of the \textit{high-quality} scenario, or $\beta=600$. }

\subsubsection{balanced-time-quality comparison of best methods}\label{sec:balance}
\textcolor{black}{This subsection provides \textit{balanced-time-quality} scenario, which is a new setting in which time and quality are in a balanced mode. Here, we compare \textit{UB-HH} with the best methods and \textit{TE-HH}. The \textit{TE-HH} has two versions. The first one is the original base heuristics (i.e., \textit{h-prob} and \textit{h-power}). Instead of the original paper setting, we have replaced the base heuristics of \textit{TE-HH} with the \textit{UB-HH} base heuristics (i.e., \textit{BS-Ex}, $k_{analytic}^{cor}$, $k_{analytic}^{uncor}$, and \textit{GCoV}) in this paper. This update enables us to compare hyper-heuristics in a fair condition. \textit{TE-HH}, in addition to the conventional beam width ($\beta$), intrinsically needs a parameter called trial beam width ($\beta_h$) to find the best base heuristic; we use the value noted in the original paper~\cite{tabataba12hyper} for $\beta_h$, which is 50. Tables~\ref{tb:randomub}--~\ref{tb:esub} show the comparison among the best methods, \textit{TE-HH}, and \textit{UB-HH}. The \textit{UB-HH} achieves the best results in five out of six benchmarks (i.e., \textit{ACO-Random}, \textit{ACO-Rat}, \textit{SARS-CoV-2}, \textit{BB}, and \textit{ES}). Concerning LCS average length for each benchmark, \textit{UB-HH} is the winner in \textit{ACO-Random}, \textit{ACO-Rat}, \textit{SARS-CoV-2}, and \textit{ES} benchmarks. For the \textit{ACO-Virus} benchmark, \textit{BS-GMPSUM} is better than other methods in both LCS average length and number of best solutions. In the \textit{BB} benchmark, \textit{TE-HH} obtains a longer average LCS length than others. As the results show, the \textit{UB-HH} generally outperforms other methods in the case of $\beta=200$.}

\begin{table}[h]
\caption{Comparison of best methods, \textit{TE-HH}, and \textit{UB-HH} over \textit{ACO-Random} benchmark in \textit{balanced-time-quality} scenario.}
\label{tb:randomub}
\centering
\resizebox{\textwidth}{!}{
\begin{tabular}{llllllll|ll}
\hline
$|\Sigma|$     & $\ell$      & n      & BS-GMPSUM &$k_{analytic}^{cor}$ &$k_{analytic}^{uncor}$    & BS-Ex  & GCoV & TE-HH & UB-HH\\ \hline
4         & 600    & 10   &218  &215 &220    & 220    & 217 & 217 & 220\\
4         & 600    & 15   &202  &201 &202    & 203    & 199 & 203 & 202\\
4         & 600    & 20   &191  &189 &192    & 191    & 187 & 191 & 192\\
4         & 600    & 25   &188  &186 &187    & 186    & 183 & 186 & 187\\
4         & 600    & 40   &173  &172 &175    & 174    & 170 & 174 & 175\\
4         & 600    & 60   &165  &164 &167    & 167    & 161 & 167 & 167\\
4         & 600    & 80   &161  &160 &162    & 161    & 157 & 162 & 162\\
4         & 600    & 100  &159  &157 &159    & 159    & 154 & 159 & 159\\
4         & 600    & 150  &151  &152 &153    & 153    & 149 & 153 & 153\\
4         & 600    & 200  &149  &149 &151    & 150    & 147 & 151 & 151\\ 
20        & 600    & 10   &61  &61 &62     & 62     & 61  & 62 & 62\\
20        & 600    & 15   &51  &50 &51     & 51     & 51  & 51 & 51\\
20        & 600    & 20   &47  &46 &47     & 47     & 47  & 47 & 47\\
20        & 600    & 25   &44  &44 &44     & 44     & 44  & 44 & 44\\
20        & 600    & 40   &38  &38 &39     & 38     & 38  & 38 & 39\\
20        & 600    & 60   &35  &34 &35     & 35     & 35  & 35 & 35\\
20        & 600    & 80   &33  &32 &33     & 33     & 33  & 33 & 33\\
20        & 600    & 100  &32  &31 &32     & 32     & 32  & 32 & 32\\
20        & 600    & 150  &29  &29 &29     & 29     & 29  & 29 & 29\\
20        & 600    & 200  &28  &28 &28     & 27     & 28  & 28 & 28\\ \hline
\multicolumn{3}{l}{Average} & 107.75 &106.9 &108.4 & 108.1 & 106.05 & 108.1 & \textbf{108.4}\\ \hline
\end{tabular}
}
\end{table}

\begin{table}[h]
\centering
\caption{Comparison of best methods, \textit{TE-HH}, and \textit{UB-HH} over \textit{ACO-Rat} benchmark in \textit{balanced-time-quality} scenario.}
\label{tb:ratub}
\resizebox{\textwidth}{!}{
\begin{tabular}{llllllll|ll}
\hline
$|\Sigma|$     & $\ell$      & n    & BS-GMPSUM & $k_{analytic}^{cor}$ &$k_{analytic}^{uncor}$    & BS-Ex  & GCoV & TE-HH & UB-HH\\ \hline
4         & 600    & 10   &198 &197 &202    & 201    & 198 & 201 & 202\\
4         & 600    & 15   &184  &181 &184    & 183    & 182 & 183 & 184\\
4         & 600    & 20   &173  &165 &172    & 171    & 169 & 172 & 169\\
4         & 600    & 25   &169  &164 &169    & 171    & 167 & 169 & 169\\
4         & 600    & 40   &155  &143 &151    & 148    & 155 & 155 & 155\\
4         & 600    & 60   &152  &150 &152    & 152    & 148 & 152 & 152\\
4         & 600    & 80   &141  &136 &139    & 140    & 141 & 141 & 141\\
4         & 600    & 100  &138  &130 &135    & 136    & 136 & 136 & 136\\
4         & 600    & 150  &124  &123 &129    & 128    & 129 & 129 & 129\\
4         & 600    & 200  &121  &119 &122    & 124    & 123 & 123 & 123\\
20        & 600    & 10   &70  &68 &70     & 70     & 70  & 70 & 70\\
20        & 600    & 15   &62  &61 &62     & 62     & 62  & 62 & 62\\
20        & 600    & 20   &54  &53 &54     & 54     & 54  & 54 & 54\\
20        & 600    & 25   &51  &50 &51     & 51     & 52  & 51 & 52\\
20        & 600    & 40   &48  &49 &49     & 48     & 49  & 49 & 49\\
20        & 600    & 60   &46  &46 &46     & 46     & 46  & 46 & 46\\
20        & 600    & 80   &43  &43 &43     & 43     & 43  & 43 & 43\\
20        & 600    & 100  &40  &38 &39     & 39     & 40  & 39 & 40\\
20        & 600    & 150  &37  &36 &37     & 37     & 37  & 37 & 37\\
20        & 600    & 200  &34  &32 &34     & 34     & 35  & 34 & 35\\ \hline
\multicolumn{3}{l}{Average} & 102 &99.2 &102 & 101.9 & 101.8 & 102.3 & \textbf{102.4}\\ \hline
\end{tabular}
}
\end{table}

\begin{table}[h]
\centering
\caption{Comparison of best methods, \textit{TE-HH}, and \textit{UB-HH} over \textit{ACO-Virus} benchmark in \textit{balanced-time-quality} scenario.}
\label{tb:virusub}
\resizebox{\textwidth}{!}{
\begin{tabular}{llllllll|ll}
\hline
$|\Sigma|$     & $\ell$      & n    & BS-GMPSUM  & $k_{analytic}^{cor}$ &$k_{analytic}^{uncor}$    & BS-Ex  & GCoV & TE-HH & UB-HH\\ \hline
4         & 600    & 10   & 225 &223 &223    & 223    & 218 & 223 & 223\\
4         & 600    & 15   & 203 &203 &203    & 202    & 204 & 202 & 203\\
4         & 600    & 20   & 191 &187 &190    & 189    & 189 & 189 & 190\\
4         & 600    & 25   & 194 &192 &193    & 193    & 190 & 193 & 193\\
4         & 600    & 40   & 170 &167 &170    & 169    & 168 & 170 & 170\\
4         & 600    & 60   & 167 &164 &166    & 166    & 162 & 166 & 166\\
4         & 600    & 80   & 161 &159 &161    & 162    & 156 & 161 & 161\\
4         & 600    & 100  & 158 &155 &156    & 157    & 154 & 157 & 156\\
4         & 600    & 150  & 152 &155 &156    & 155    & 152 & 156 & 156\\
4         & 600    & 200  & 149 &153 &155    & 153    & 149 & 153 & 155\\
20        & 600    & 10   & 74 &74 &74     & 74     & 75  & 75 & 75\\
20        & 600    & 15   & 62 &62 &63     & 64     & 62  & 63 & 62\\
20        & 600    & 20   & 59 &58 &59     & 59     & 59  & 59 & 59\\
20        & 600    & 25   & 55 &54 &55     & 54     & 55  & 55 & 55\\
20        & 600    & 40   & 51 &49 &50     & 49     & 49  & 50 & 49\\
20        & 600    & 60   & 48 &47 &48     & 48     & 47  & 47 & 47\\
20        & 600    & 80   & 46 &46 &46     & 46     & 45  & 46 & 45\\
20        & 600    & 100  & 44 &44 &44     & 43     & 44  & 43 & 44\\
20        & 600    & 150  & 45 &45 &45     & 45     & 45  & 45 & 45\\
20        & 600    & 200  & 44 &44 &43     & 44     & 43  & 43 & 43\\ \hline
\multicolumn{3}{l}{Average} & 114.9 &114.05 &\textbf{115} & 114.75 & 113.3 & 114.8 & 114.85\\ \hline
\end{tabular}
}
\end{table}

\begin{table}[h]
\caption{Comparison of best methods, \textit{TE-HH}, and \textit{UB-HH} over \textit{SARS-CoV-2} benchmark in \textit{balanced-time-quality} scenario.}
\centering
\label{tb:sarsub}
\resizebox{\textwidth}{!}{
\begin{tabular}{llllllll|ll}
\hline
$|\Sigma|$     & $\ell$      & n    & BS-GMPSUM  & $k_{analytic}^{cor}$ &$k_{analytic}^{uncor}$    & BS-Ex  & GCoV & TE-HH & UB-HH\\ \hline
4&400&10 & 189 &  198&179 & 182 & 172 & 179 & 198 \\
4&400&20 & 193 &  191&195 & 189 & 176 & 191 & 191 \\
4&400&30 & 178 &  175&174 & 173 & 156 & 175 & 175 \\
4&400&40 & 164 &  168&154 & 150 & 146 & 168 & 168 \\
4&400&50 & 157 &  160&156 & 145 & 141 & 160 &  160\\
4&400&60 & 153 &  153&146 & 139 & 139 & 153 &  153\\
4&400&70 & 148 &  142&140 & 137 & 136 & 140 &  142\\
4&400&80 & 151 &  150&147 & 140 & 140 & 150 &  150\\
4&400&90 & 158 &  163&152 & 150 & 145 & 163 &  163\\
4&400&100& 144 &  148&140 & 139 & 132 & 148 &  148\\
4&400&110& 141 &  143&142 & 143 & 133 & 143 &  143\\
\hline
\multicolumn{3}{l}{Average} &161.45 & 162.81 &156.81 & 153.4 & 146.9 & 160.9 & \textbf{162.81} \\ \hline
\end{tabular}
}
\end{table}

\begin{table}[h]
\caption{Comparison of best methods, \textit{TE-HH}, and \textit{UB-HH} over \textit{BB} benchmark in \textit{balanced-time-quality} scenario.}
\label{tb:bbub}
\centering
\resizebox{\textwidth}{!}{
\begin{tabular}{llllllll|ll}
\hline
$|\Sigma|$     & $\ell$      & n     & BS-GMPSUM & $k_{analytic}^{cor}$ &$k_{analytic}^{uncor}$    & BS-Ex  & GCoV & TE-HH & UB-HH\\ \hline
2  &   1000    &    10  & 629  &  635.7& 594.1 & 604.8 & 614.1 & 631.5 & 635.7 \\
2  &   1000    &    100 & 556  &  559.4& 540.3 & 531.3 & 543.1 & 559.4 & 559.4 \\
4  &   1000    &    10  & 467.5  &  464.3& 426.9 & 424.4 & 452.4 & 463.4 & 464.3 \\
4  &   1000    &    100 & 364.5  &  365.6& 341.8 & 324.7 & 365.3 & 366.6 & 365.6 \\
8  &   1000    &    10  & 343.4  &  359.6& 307.2 & 289.9 & 359.5 & 368.3 & 359.6 \\
8  &   1000    &    100 & 244.1  &  242.1& 217 & 201.5 & 244.3 & 242.5 & 242.1 \\
24  &  1000    &    10  & 263  &  274.1& 226.7 & 220.6 & 277.1 & 296.7 & 274.1 \\
24  &  1000    &    100 & 128.9  &  130.8& 112.4 & 102.2 & 123   & 127.5 & 130.8 \\
\hline
\multicolumn{3}{l}{Average} & 374.63 & 378.95& 345.8 & 337.42 & 372.35 & \textbf{381.98} & 378.95 \\ \hline
\end{tabular}
}
\end{table}

\begin{table}[h]
\caption{Comparison of best methods, \textit{TE-HH}, and \textit{UB-HH} over \textit{ES} benchmark in \textit{balanced-time-quality} scenario.}
\label{tb:esub}
\centering
\resizebox{\textwidth}{!}{
\begin{tabular}{llllllll|ll}
\hline
$|\Sigma|$     & $\ell$      & n    & BS-GMPSUM  & $k_{analytic}^{cor}$ &$k_{analytic}^{uncor}$    & BS-Ex  & GCoV & TE-HH & UB-HH\\ \hline
2        & 1000    & 10     &603.72 & 600.8 & 584.84 & 609.96 & 527.28 & 609.96 & 609.96\\
2        & 1000    & 50     &353.98 & 532.66 & 531.62 & 537.02 & 510.8  & 537.02 & 536.9 \\
2        & 1000    & 100    &518.12 & 515.28 & 519    & 518.96 & 500.2  & 518.84 & 518.96\\
10       & 1000    & 10     &197.44 & 195.32 & 198.88 & 199.16 & 190.13 & 198.86 & 198.88\\
10       & 1000    & 50     &135.34 & 134.38 & 135.68 & 135.66 & 127.53 & 135.48 & 135.68\\
10       & 1000    & 100    &122.74 & 121.72 & 122.82 & 122.7  & 116.35 & 122.62 & 122.82\\
25       & 2500    & 10     &229.04 & 226.4 & 229.9  & 226.06 & 214.06 & 229.8  & 229.9 \\
25       & 2500    & 50     &138.61 & 136.98 & 138.74 & 138.7  & 128.92 & 138.66 & 138.74\\
25       & 2500    & 100    &122.21 & 121.04 & 122.3  & 122.38 & 113.33 & 122.22 & 122.3 \\
100      & 5000    & 10     &138.98 & 137.36 & 139.08 & 138.94 & 134.81 & 138.98 & 139.08\\
100      & 5000    & 50     &70.9 & 69.84 & 71     & 70.98  & 66.3   & 70.98  & 71    \\
100      & 5000    & 100    &59.93 & 59.2 & 60.06  & 59.98  & 56.74  & 60.04  & 60.06 \\ \hline 
\multicolumn{3}{l}{Average} &239.41 & 237.58  & 237.82 & 240.04 & 227.62 & 240.028 & \textbf{240.35}\\ \hline
\end{tabular}
}
\end{table}

\subsubsection{low-time comparison of best methods}\label{sec:low}
\textcolor{black}{We set the $\beta$ for all methods to $50$ in the second scenario --- \textit{low-time} scenario. Tables~\ref{tb:random50}--~\ref{tb:es50} show the detailed comparison of \textit{UB-HH} with the best methods in the \textit{low-time} scenario. The \textit{UB-HH} obtains the highest LCS average length and number of best solutions in four of six benchmarks, i.e., \textit{ACO-Random}, \textit{ACO-Rat}, \textit{BB}, and \textit{ES}. In the case of the \textit{ACO-Virus} dataset, $k_{analytic}^{uncor}$ could obtain the best average rank and the highest number of best solutions. For the \textit{SARS-CoV-2} dataset, \textit{BS-GMPSUM} is the winner in both LCS average length and number of best solutions. To sum up, the lowest average rank belongs to  \textit{UB-HH} in general in the case of $\beta=50$.}

\begin{table}[h]
\caption{\textcolor{black}{Comparison of best methods and \textit{UB-HH} over \textit{ACO-Random} benchmark in \textit{low-time} scenario.}}
\label{tb:random50}
\centering
\resizebox{\textwidth}{!}{
\begin{tabular}{lllllllll}
\hline
$|\Sigma|$     & $\ell$      & n      & BS-GMPSUM &$k_{analytic}^{cor}$ &$k_{analytic}^{uncor}$    & BS-Ex  & GCoV & UB-HH\\ \hline
4         & 600    & 10 &211 &214 &215 &215 &\textbf{217} &215   \\
4         & 600    & 15 &196 &197 &\textbf{201} &\textbf{201} &196 &\textbf{201}   \\
4         & 600    & 20 &189 &187 &\textbf{191} &\textbf{191} &187 &\textbf{191}   \\
4         & 600    & 25 &184 &182 &184 &\textbf{186} &181 &184   \\
4         & 600    & 40 &171 &170 &\textbf{173} &\textbf{173} &168 &\textbf{173}   \\
4         & 600    & 60 &164 &163 &\textbf{166} &\textbf{166} &159 &\textbf{166}   \\
4         & 600    & 80 &159 &158 &\textbf{161} &\textbf{161} &154 &\textbf{161}   \\
4         & 600    & 100&157 &156 &\textbf{158} &\textbf{158} &152 &\textbf{158}   \\
4         & 600    & 150&150 &151 &\textbf{152} &\textbf{152} &148 &\textbf{152}   \\
4         & 600    & 200&149 &148 &\textbf{151} &150 &147 &\textbf{151}   \\ 
20        & 600    & 10 &59 &58 &61 &61 &61 &61   \\
20        & 600    & 15 &50 &50 &\textbf{51} &50 &50 &\textbf{51}   \\
20        & 600    & 20 &46 &43 &46 &\textbf{47} &46 &46   \\
20        & 600    & 25 &43 &43 &43 &43 &43 &43   \\
20        & 600    & 40 &37 &38 &38 &38 &38 &38   \\
20        & 600    & 60 &34 &34 &34 &34 &34 &34   \\
20        & 600    & 80 &32 &32 &\textbf{33} &32 &32 &\textbf{33}   \\
20        & 600    & 100&31 &31 &31 &31 &31 &31   \\
20        & 600    & 150&29 &29 &29 &29 &29 &29   \\
20        & 600    & 200&28 &28 &28 &27 &28 &28   \\ \hline
\multicolumn{3}{l}{Average} &105.95 &105.6 &\textbf{107.3} &107.25 &105.05 & \textbf{107.3} \\ \hline
\end{tabular}
}
\end{table}

\begin{table}[h]
\centering
\caption{\textcolor{black}{Comparison of best methods and \textit{UB-HH} over \textit{ACO-Rat} benchmark in \textit{low-time} scenario.}}
\label{tb:rat50}
\resizebox{\textwidth}{!}{
\begin{tabular}{lllllllll}
\hline
$|\Sigma|$     & $\ell$      & n    & BS-GMPSUM & $k_{analytic}^{cor}$ &$k_{analytic}^{uncor}$    & BS-Ex  & GCoV & UB-HH\\ \hline
4         & 600    & 10  &197 &192 &\textbf{198} &\textbf{198} &196 &\textbf{198}   \\
4         & 600    & 15  &179 &179 &180 &\textbf{182} &180 &180   \\
4         & 600    & 20  &168 &165 &\textbf{170} &169 &169 &169   \\
4         & 600    & 25  &163 &161 &\textbf{167} &166 &166 &\textbf{167}   \\
4         & 600    & 40  &144 &137 &144 &143 &\textbf{150} &\textbf{150}   \\
4         & 600    & 60  &148 &147 &\textbf{152} &150 &148 &\textbf{152}   \\
4         & 600    & 80  &136 &136 &139 &138 &\textbf{140} &\textbf{140}   \\
4         & 600    & 100 &128 &129 &134 &\textbf{136} &134 &134   \\
4         & 600    & 150 &123 &121 &\textbf{129} &127 &127 &127   \\
4         & 600    & 200 &120 &119 &118 &121 &\textbf{124} &\textbf{124}   \\
20        & 600    & 10  &68 &68 &69 &\textbf{70} &69 &69   \\
20        & 600    & 15  &\textbf{62} &59 &\textbf{62} &\textbf{62} &60 &60   \\
20        & 600    & 20  &\textbf{53} &52 &52 &52 &52 &52   \\
20        & 600    & 25  &51 &50 &50 &51 &51 &51   \\
20        & 600    & 40  &48 &48 &47 &47 &\textbf{49} &\textbf{49}   \\
20        & 600    & 60  &46 &46 &46 &45 &\textbf{47} &\textbf{47}   \\
20        & 600    & 80  &\textbf{43} &42 &42 &41 &\textbf{43} &\textbf{43}   \\
20        & 600    & 100 &38 &37 &38 &38 &38 &38   \\
20        & 600    & 150 &36 &35 &\textbf{37} &36 &\textbf{37} &\textbf{37}   \\
20        & 600    & 200 &33 &32 &33 &\textbf{34} &33 &33   \\ \hline
\multicolumn{3}{l}{Average} &99.2 &97.75 &100.35 &100.3 &100.65 &\textbf{101} \\ \hline
\end{tabular}
}
\end{table}

\begin{table}[h]
\centering
\caption{\textcolor{black}{Comparison of best methods and \textit{UB-HH} over \textit{ACO-Virus} benchmark in \textit{low-time} scenario.}}
\label{tb:virus50}
\resizebox{\textwidth}{!}{
\begin{tabular}{lllllllll}
\hline
$|\Sigma|$     & $\ell$      & n    & BS-GMPSUM  & $k_{analytic}^{cor}$ &$k_{analytic}^{uncor}$    & BS-Ex  & GCoV & UB-HH\\ \hline
4         & 600    & 10  &221 &220 &\textbf{223} &\textbf{223} &218 &\textbf{223} \\
4         & 600    & 15  &198 &199 &\textbf{201} &\textbf{201} &199 &\textbf{201} \\
4         & 600    & 20  &187 &186 &187 &\textbf{189} &\textbf{189} &187 \\
4         & 600    & 25  &\textbf{191} &\textbf{191} &190 &190 &188 &190 \\
4         & 600    & 40  &167 &165 &\textbf{168} &167 &165 &\textbf{168} \\
4         & 600    & 60  &162 &162 &\textbf{163} &162 &161 &\textbf{163} \\
4         & 600    & 80  &157 &156 &157 &\textbf{158} &155 &157 \\
4         & 600    & 100 &153 &152 &155 &\textbf{156} &154 &155 \\
4         & 600    & 150 &154 &\textbf{155} &\textbf{155} &154 &150 &\textbf{155} \\
4         & 600    & 200 &153 &153 &153 &152 &149 &153 \\
20        & 600    & 10  &74 &72 &73 &73 &\textbf{75} &\textbf{75} \\
20        & 600    & 15  &61 &61 &\textbf{63} &\textbf{63} &60 &60 \\
20        & 600    & 20  &58 &58 &59 &59 &59 &59 \\
20        & 600    & 25  &55 &53 &55 &54 &55 &55 \\
20        & 600    & 40  &49 &49 &\textbf{50} &49 &49 &49 \\
20        & 600    & 60  &46 &46 &46 &46 &46 &46 \\
20        & 600    & 80  &45 &44 &\textbf{46} &45 &45 &45 \\
20        & 600    & 100 &44 &44 &44 &44 &43 &43 \\
20        & 600    & 150 &45 &45 &45 &45 &44 &44 \\
20        & 600    & 200 &43 &43 &43 &42 &43 &43 \\ \hline
\multicolumn{3}{l}{Average} &113.15 &112.7 &\textbf{113.8} &113.6 &112.35 &113.55 \\ \hline
\end{tabular}
}
\end{table}

\begin{table}[h]
\caption{\textcolor{black}{Comparison of best methods and \textit{UB-HH} over \textit{SARS-CoV-2} benchmark in \textit{low-time} scenario.}}
\centering
\label{tb:sars50}
\resizebox{\textwidth}{!}{
\begin{tabular}{lllllllll}
\hline
$|\Sigma|$     & $\ell$      & n    & BS-GMPSUM  & $k_{analytic}^{cor}$ &$k_{analytic}^{uncor}$    & BS-Ex  & GCoV & UB-HH\\ \hline
4&400&10   &189 &192 &\textbf{195} &186 &168 &192     \\
4&400&20   &\textbf{195} &190 &186 &183 &175 &190     \\
4&400&30   &\textbf{190} &176 &167 &160 &156 &176     \\
4&400&40   &154 &\textbf{157} &151 &144 &146 &\textbf{157}     \\
4&400&50   &\textbf{153} &152 &151 &148 &143 &152     \\
4&400&60   &147 &\textbf{149} &143 &142 &136 &\textbf{149}     \\
4&400&70   &139 &139 &\textbf{143} &134 &133 &139     \\
4&400&80   &\textbf{148} &146 &146 &139 &135 &146     \\
4&400&90   &\textbf{155} &\textbf{155} &152 &152 &137 &\textbf{155}     \\
4&400&100  &\textbf{141} &140 &139 &139 &129 &140     \\
4&400&110  &\textbf{144} &143 &140 &135 &127 &143     \\ \hline
\multicolumn{3}{l}{Average} &\textbf{159.54} &158.09 &155.72 &151.09 &144.09 &158.09 \\ \hline
\end{tabular}
}
\end{table}

\begin{table}[h]
\caption{\textcolor{black}{Comparison of best methods and \textit{UB-HH} over \textit{BB} benchmark in \textit{low-time} scenario.}}
\label{tb:bb50}
\centering
\resizebox{\textwidth}{!}{
\begin{tabular}{lllllllll}
\hline
$|\Sigma|$     & $\ell$      & n     & BS-GMPSUM & $k_{analytic}^{cor}$ &$k_{analytic}^{uncor}$    & BS-Ex  & GCoV & UB-HH\\ \hline
2  &   1000    &    10  &\textbf{627.6} &625.6 &587.1 &607.6 &613.1 &625.6  \\
2  &   1000    &    100 &551.4 &\textbf{553.7} &532.5 &525.2 &539.3 &\textbf{553.7}  \\
4  &   1000    &    10  &\textbf{462.3} &454.2 &420.2 &413.3 &451.2 &454.2  \\
4  &   1000    &    100 &359   &357.7 &338.1 &319   &\textbf{359.7} &357.7  \\
8  &   1000    &    10  &361.5 &366.3 &306.4 &294.2 &\textbf{367.2} &366.3  \\
8  &   1000    &    100 &233.6 &231.5 &210.8 &195.6 &\textbf{240.5} &231.5  \\
24  &  1000    &    10  &262.8 &\textbf{274.6} &240.1 &204.3 &265.7 &\textbf{274.6}  \\
24  &  1000    &    100 &121.3 &\textbf{125}   &103.2 &96.9  &123.6 &\textbf{125}  \\ \hline
\multicolumn{3}{l}{Average} &372.43 &\textbf{373.57} &342.3 &332.01 &370.03 &\textbf{373.57}  \\ \hline
\end{tabular}
}
\end{table}

\begin{table}[h]
\caption{\textcolor{black}{Comparison of best methods and \textit{UB-HH} over \textit{ES} benchmark in \textit{low-time} scenario.}}
\label{tb:es50}
\centering
\resizebox{\textwidth}{!}{
\begin{tabular}{lllllllll}
\hline
$|\Sigma|$     & $\ell$      & n    & BS-GMPSUM  & $k_{analytic}^{cor}$ &$k_{analytic}^{uncor}$    & BS-Ex  & GCoV & UB-HH\\ \hline
2        & 1000    & 10   &600.92 &597.98 &582.56 &\textbf{607.22} &584.2  &\textbf{607.22}   \\
2        & 1000    & 50   &533.6  &530.46 &529.6  &\textbf{535.04} &513    &\textbf{535.04}   \\
2        & 1000    & 100  &516.16 &513.6  &\textbf{517.52} &517.34 &502.6  &517.34   \\
10       & 1000    & 10   &195.18 &194.16 &196.84 &\textbf{197.18} &193.1  &196.84   \\
10       & 1000    & 50   &134.28 &133.04 &\textbf{134.62} &134.48 &129    &\textbf{134.62}   \\
10       & 1000    & 100  &121.94 &120.86 &\textbf{122.04} &121.94 &117.8  &\textbf{122.04}   \\
25       & 2500    & 10   &225.6  &224.22 &\textbf{227.76} &221.2  &220.4  &\textbf{227.76}   \\
25       & 2500    & 50   &137.32 &135.82 &137.64 &\textbf{137.82} &130.4  &137.64   \\
25       & 2500    & 100  &121.6  &120.2  &121.6  &\textbf{121.74} &114.7  &121.6    \\
100      & 5000    & 10   &135.64 &135.34 &\textbf{137.72} &135.64 &132.1  &\textbf{137.72}   \\
100      & 5000    & 50   &70.21  &69.14  &70.3   &\textbf{70.38}  &67.8   &70.3     \\
100      & 5000    & 100  &59.33  &58.76  &\textbf{59.6}   &\textbf{59.6}   &58.1   &\textbf{59.6}     \\ \hline 
\multicolumn{3}{l}{Average} &237.64 &236.13 &236.48 &238.29 &230.27 &\textbf{238.97} \\ \hline
\end{tabular}
}
\end{table}

\subsubsection{high-quality comparison of best methods}\label{sec:high}
\textcolor{black}{In the third scenario, we compare the \textit{UB-HH} algorithm with the best methods in the \textit{high-quality} setting, where it is $\beta=600$. Tables~\ref{tb:random600}--~\ref{tb:es600} show that the \textit{UB-HH} obtains a higher number of best solutions in four out of six benchmarks (i.e., \textit{ACO-Rat}, \textit{SARS-CoV-2}, \textit{BB}, and \textit{ES}) and the best LCS average length in \textit{ACO-Rat}, \textit{BB}, and \textit{ES} benchmarks. \textit{BS-Ex} obtains a better LCS average length and more best solutions concerning the \textit{ACO-Random} benchmark. For \textit{ACO-Virus}, \textit{BS-GMPSUM} is better than other methods in both LCS average length and number of best solutions. However, concerning the lowest average rank, \textit{UB-HH} is better than all methods overall.}

\begin{table}[h]
\caption{\textcolor{black}{Comparison of best methods and \textit{UB-HH} over \textit{ACO-Random} benchmark in \textit{high-quality} scenario.}}
\label{tb:random600}
\centering
\resizebox{\textwidth}{!}{
\begin{tabular}{lllllllll}
\hline
$|\Sigma|$     & $\ell$      & n      & BS-GMPSUM &$k_{analytic}^{cor}$ &$k_{analytic}^{uncor}$    & BS-Ex  & GCoV & UB-HH\\ \hline
4         & 600    & 10  &218 &215 &218 &\textbf{219} &216 &218   \\
4         & 600    & 15  &204 &203 &204 &204 &201 &204   \\
4         & 600    & 20  &192 &191 &\textbf{193} &\textbf{193} &190 &\textbf{193}   \\
4         & 600    & 25  &187 &186 &187 &187 &184 &187   \\
4         & 600    & 40  &174 &172 &\textbf{175} &\textbf{175} &170 &\textbf{175}   \\
4         & 600    & 60  &167 &166 &\textbf{168} &\textbf{168} &162 &\textbf{168}   \\
4         & 600    & 80  &162 &160 &162 &\textbf{163} &159 &162   \\
4         & 600    & 100 &159 &158 &159 &159 &155 &159   \\
4         & 600    & 150 &152 &152 &\textbf{153} &\textbf{153} &149 &\textbf{153}   \\
4         & 600    & 200 &150 &149 &\textbf{151} &\textbf{151} &148 &\textbf{151}   \\ 
20        & 600    & 10  &62 &61 &62 &62 &62 &62  \\
20        & 600    & 15  &\textbf{52} &51 &51 &51 &51 &51  \\
20        & 600    & 20  &47 &47 &47 &47 &47 &47  \\
20        & 600    & 25  &44 &44 &44 &44 &44 &44  \\
20        & 600    & 40  &39 &38 &39 &39 &38 &39  \\
20        & 600    & 60  &35 &34 &35 &35 &35 &35  \\
20        & 600    & 80  &33 &33 &33 &33 &33 &33  \\
20        & 600    & 100 &32 &31 &32 &32 &31 &32  \\
20        & 600    & 150 &29 &29 &29 &29 &29 &29  \\
20        & 600    & 200 &28 &28 &28 &27 &28 &28  \\ \hline
\multicolumn{3}{l}{Average} &108.3 &107.4 &108.5 &\textbf{108.55} &106.6 & 108.5 \\ \hline
\end{tabular}
}
\end{table}

\begin{table}[h]
\centering
\caption{\textcolor{black}{Comparison of best methods and \textit{UB-HH} over \textit{ACO-Rat} benchmark in \textit{high-quality} scenario.}}
\label{tb:rat600}
\resizebox{\textwidth}{!}{
\begin{tabular}{lllllllll}
\hline
$|\Sigma|$     & $\ell$      & n    & BS-GMPSUM & $k_{analytic}^{cor}$ &$k_{analytic}^{uncor}$    & BS-Ex  & GCoV & UB-HH\\ \hline
4         & 600    & 10  &200  &198  &\textbf{204}  &\textbf{204}  &202  &\textbf{204}   \\
4         & 600    & 15  &183  &183  &\textbf{185}  &\textbf{185}  &184  &\textbf{185}   \\
4         & 600    & 20  &\textbf{173}  &167  &\textbf{173}  &172  &171  &171   \\
4         & 600    & 25  &169  &165  &\textbf{170}  &\textbf{170}  &168  &\textbf{170}   \\
4         & 600    & 40  &\textbf{155}  &145  &148  &152  &153  &153   \\
4         & 600    & 60  &\textbf{153}  &149  &152  &152  &149  &152   \\
4         & 600    & 80  &141  &137  &\textbf{142}  &\textbf{142}  &140  &140   \\
4         & 600    & 100 &137  &134  &135  &137  &137  &137   \\
4         & 600    & 150 &125  &124  &128  &\textbf{129}  &\textbf{129}  &\textbf{129}   \\
4         & 600    & 200 &119  &119  &123  &\textbf{124}  &\textbf{124}  &\textbf{124}   \\
20        & 600    & 10  &70  &70  &70  &70  &70  &70   \\
20        & 600    & 15  &62  &62  &62  &\textbf{63}  &62  &62   \\
20        & 600    & 20  &54  &53  &54  &54  &54  &54   \\
20        & 600    & 25  &51  &50  &51  &51  &\textbf{52}  &\textbf{52}   \\
20        & 600    & 40  &48  &49  &49  &49  &\textbf{50}  &\textbf{50}   \\
20        & 600    & 60  &47  &46  &47  &46  &\textbf{48}  &\textbf{48}   \\
20        & 600    & 80  &43  &\textbf{44}  &43  &43  &\textbf{44}  &\textbf{44}   \\
20        & 600    & 100 &40  &39  &39  &40  &40  &40   \\
20        & 600    & 150 &38  &36  &38  &37  &38  &38   \\
20        & 600    & 200 &35  &33  &35  &34  &35  &35   \\ \hline
\multicolumn{3}{l}{Average} &102.15 &100.15 &102.4 &102.7 &102.5 &\textbf{102.9} \\ \hline
\end{tabular}
}
\end{table}

\begin{table}[h]
\centering
\caption{\textcolor{black}{Comparison of best methods and \textit{UB-HH} over \textit{ACO-Virus} benchmark in \textit{high-quality} scenario.}}
\label{tb:virus600}
\resizebox{\textwidth}{!}{
\begin{tabular}{lllllllll}
\hline
$|\Sigma|$     & $\ell$      & n    & BS-GMPSUM  & $k_{analytic}^{cor}$ &$k_{analytic}^{uncor}$    & BS-Ex  & GCoV & UB-HH\\ \hline
4         & 600    & 10  &\textbf{225} &224 &223 &223 &222 &223 \\
4         & 600    & 15  &203 &201 &204 &\textbf{205} &202 &204 \\
4         & 600    & 20  &191 &187 &190 &\textbf{192} &191 &190 \\
4         & 600    & 25  &\textbf{195} &193 &194 &194 &193 &194 \\
4         & 600    & 40  &\textbf{171} &168 &\textbf{171} &170 &168 &\textbf{171} \\
4         & 600    & 60  &\textbf{167} &164 &166 &166 &165 &166 \\
4         & 600    & 80  &162 &159 &160 &\textbf{163} &158 &160 \\
4         & 600    & 100 &\textbf{160} &158 &159 &158 &155 &159 \\
4         & 600    & 150 &156 &156 &\textbf{157} &156 &152 &\textbf{157} \\
4         & 600    & 200 &150 &154 &\textbf{155} &154 &151 &\textbf{155} \\
20        & 600    & 10  &\textbf{76} &74 &75 &\textbf{76} &75 &75 \\
20        & 600    & 15  &63 &63 &\textbf{64} &\textbf{64} &63 &63 \\
20        & 600    & 20  &59 &59 &59 &59 &\textbf{60} &\textbf{60} \\
20        & 600    & 25  &55 &54 &55 &54 &55 &55 \\
20        & 600    & 40  &\textbf{51} &49 &50 &50 &\textbf{51} &\textbf{51} \\
20        & 600    & 60  &48 &47 &48 &48 &48 &48 \\
20        & 600    & 80  &46 &46 &46 &46 &46 &46 \\
20        & 600    & 100 &\textbf{45} &44 &\textbf{45} &44 &44 &44 \\
20        & 600    & 150 &45 &45 &45 &45 &45 &45 \\
20        & 600    & 200 &44 &44 &43 &43 &44 &44 \\ \hline
\multicolumn{3}{l}{Average} &\textbf{115.6} &114.45 &115.45 &115.5 &114.4 &115.5 \\ \hline
\end{tabular}
}
\end{table}

\begin{table}[h]
\caption{\textcolor{black}{Comparison of best methods and \textit{UB-HH} over \textit{SARS-CoV-2} benchmark in \textit{high-quality} scenario.}}
\centering
\label{tb:sars600}
\resizebox{\textwidth}{!}{
\begin{tabular}{lllllllll}
\hline
$|\Sigma|$     & $\ell$      & n    & BS-GMPSUM  & $k_{analytic}^{cor}$ &$k_{analytic}^{uncor}$    & BS-Ex  & GCoV & UB-HH\\ \hline
4&400&10   &196 &198 &195 &\textbf{201} &168 &198     \\
4&400&20   &\textbf{224} &\textbf{224} &187 &189 &176 &\textbf{224}     \\
4&400&30   &183 &173 &\textbf{187} &174 &166 &173     \\
4&400&40   &164 &\textbf{169} &156 &153 &146 &\textbf{169}     \\
4&400&50   &\textbf{161} &160 &156 &156 &146 &160     \\
4&400&60   &155 &\textbf{156} &153 &150 &143 &\textbf{156}     \\
4&400&70   &\textbf{149} &143 &143 &139 &141 &143     \\
4&400&80   &\textbf{158} &153 &150 &146 &138 &153     \\
4&400&90   &\textbf{163} &161 &154 &150 &139 &161     \\
4&400&100  &144 &\textbf{146} &143 &142 &136 &\textbf{146}     \\
4&400&110  &149 &\textbf{150} &145 &144 &134 &\textbf{150}     \\ \hline
\multicolumn{3}{l}{Average} &\textbf{167.81} &166.63 &160.81 &158.54 &148.45 &166.63 \\ \hline
\end{tabular}
}
\end{table}

\begin{table}[h]
\caption{\textcolor{black}{Comparison of best methods and \textit{UB-HH} over \textit{BB} benchmark in \textit{high-quality} scenario.}}
\label{tb:bb600}
\centering
\resizebox{\textwidth}{!}{
\begin{tabular}{lllllllll}
\hline
$|\Sigma|$     & $\ell$      & n     & BS-GMPSUM & $k_{analytic}^{cor}$ &$k_{analytic}^{uncor}$    & BS-Ex  & GCoV & UB-HH\\ \hline
2  &   1000    &    10  &632.8 &\textbf{634.9} &587.4 &614.6 &620.9 &\textbf{634.9}  \\
2  &   1000    &    100 &560.9 &\textbf{562}   &541.8 &536.3 &544.7 &\textbf{562}  \\
4  &   1000    &    10  &\textbf{468.2} &453   &421   &419.2 &459.8 &453  \\
4  &   1000    &    100 &366.8 &\textbf{372.3} &345.9 &328.1 &366.4 &\textbf{372.3}  \\
8  &   1000    &    10  &353.9 &341.4 &317.6 &303.6 &\textbf{363.1} &341.4  \\
8  &   1000    &    100 &246.1 &\textbf{247.9} &222.6 &209.4 &242.8 &\textbf{247.9}  \\
24  &  1000    &    10  &265.1 &\textbf{281.8} &233   &225.6 &272.5 &\textbf{281.8}  \\
24  &  1000    &    100 &130.7 &\textbf{134.1} &114.7 &109.5 &127.8 &\textbf{134.1}  \\ \hline
\multicolumn{3}{l}{Average} &378.06 &\textbf{378.42} &348 &343.28 &374.75 &\textbf{378.42}  \\ \hline
\end{tabular}
}
\end{table}

\begin{table}[h]
\caption{\textcolor{black}{Comparison of best methods and \textit{UB-HH} over \textit{ES} benchmark in \textit{high-quality} scenario.}}
\label{tb:es600}
\centering
\resizebox{\textwidth}{!}{
\begin{tabular}{lllllllll}
\hline
$|\Sigma|$     & $\ell$      & n    & BS-GMPSUM  & $k_{analytic}^{cor}$ &$k_{analytic}^{uncor}$    & BS-Ex  & GCoV & UB-HH\\ \hline
2        & 1000    & 10   &605.94 &603    &585.94 &\textbf{611.48} &593.52 &\textbf{611.48}   \\
2        & 1000    & 50   &537.08 &533.94 &532.6  &\textbf{538.22} &517.7  &\textbf{538.22}   \\
2        & 1000    & 100  &519.28 &516.26 &\textbf{519.92} &519.84 &503.62 &519.84   \\
10       & 1000    & 10   &198.72 &197.42 &200.6  &\textbf{200.68} &199.1  &200.6   \\
10       & 1000    & 50   &136.19 &134.96 &\textbf{136.22} &136.14 &132.1  &\textbf{136.22}   \\
10       & 1000    & 100  &122.76 &122.18 &\textbf{123.28} &123.22 &119.6  &\textbf{123.28}   \\
25       & 2500    & 10   &229.9  &228.76 &\textbf{231.92} &228.5  &221.2  &\textbf{231.92}   \\
25       & 2500    & 50   &139.23 &137.66 &\textbf{139.38}  &139.34 &133.4  &\textbf{139.38}   \\
25       & 2500    & 100  &122.65 &121.44 &\textbf{122.78} &122.74 &119.71 &\textbf{122.78}   \\
100      & 5000    & 10   &140.75 &139.45 &140.78 &\textbf{141.14} &137.1  &140.78   \\
100      & 5000    & 50   &71.27  &70.1   &71.28  &\textbf{71.32}  &67.5   &71.28   \\
100      & 5000    & 100  &60.08  &59.84  &60.12  &\textbf{60.16}  &56.94  &60.12   \\ \hline 
\multicolumn{3}{l}{Average} &240.32 &238.75 &238.73 &241.06 &233.45 &\textbf{241.32} \\ \hline
\end{tabular}
}
\end{table}

\subsection{Time comparison}\label{sec:time}
\begin{figure}
    \centering
    \includegraphics[width=0.8\columnwidth]{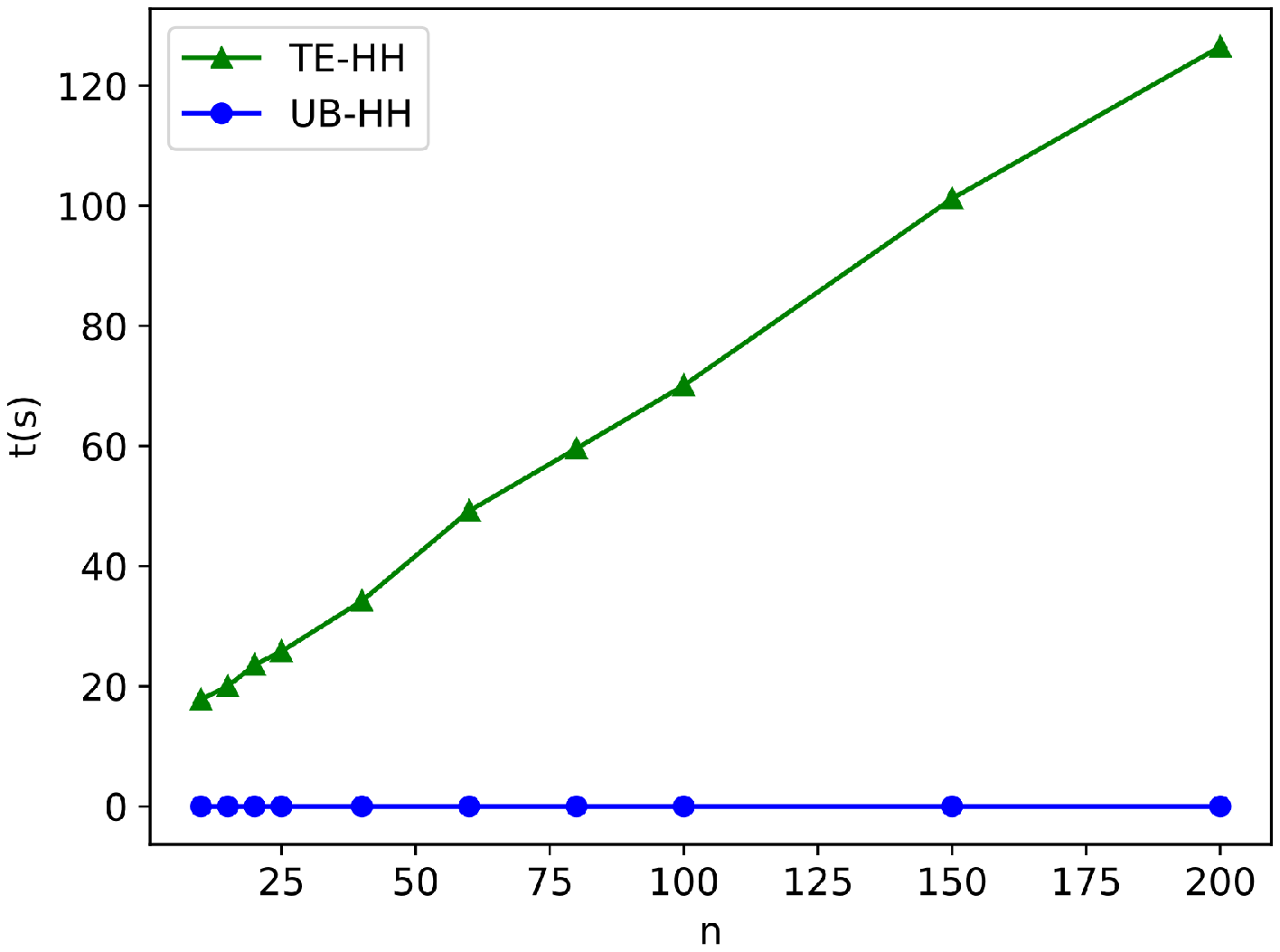}
    \caption{Comparison of the run time of \textit{TE-HH} and \textit{UB-HH} on \textit{ACO-Random} dataset with $|\Sigma| = 4$. \textcolor{black}{The same pattern happens for other datasets.}}
    \label{fig:time}
\end{figure}

\textcolor{black}{Here, we compare the run time aspect of \textit{TE-HH} and \textit{UB-HH} to choose the suitable heuristic among the base heuristics. \textit{TE-HH} needs to run all its base heuristics with $\beta_h$. So, by having four base heuristics, \textit{TE-HH} run time is the summation of the running time of \textit{BS-Ex}, $k_{analytic}^{cor}$, $k_{analytic}^{uncor}$, and \textit{GCoV} with $\beta_h$. In contrast, \textit{UB-HH} does not need to run base heuristics, and it instantly finds the suitable base heuristic based on strings' internal properties (less than one second). Among six datasets, Fig.~\ref{fig:time} shows the run time of \textit{TE-HH} and \textit{UB-HH} on the \textit{ACO-Random} dataset. As the figure shows, \textit{UB-HH} can instantly specify the suitable heuristic, but \textit{TE-HH} run time depends on the run time of all heuristics and is very high compared to \textit{UB-HH}. The same pattern happens for the other five benchmarks i.e., \textit{ACO-Rat}, \textit{ACO-Virus}, \textit{SARS-CoV-2}, \textit{BB}, and \textit{ES}.}

\section{Statistical analysis}\label{sec:statsign}
\begin{figure}[H]
     \centering
     \begin{subfigure}[]{0.45\textwidth}
         \centering
         \includegraphics[width=\textwidth]{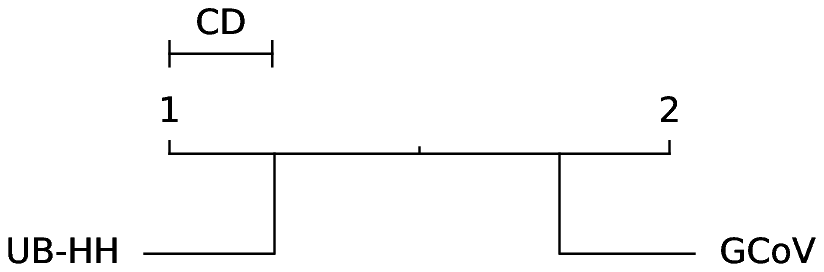}
         \caption{\textit{GCoV} and \textit{UB-HH}}
         \label{fig:gcov200}
     \end{subfigure}
     \begin{subfigure}[]{0.45\textwidth}
         \centering
         \includegraphics[width=\textwidth]{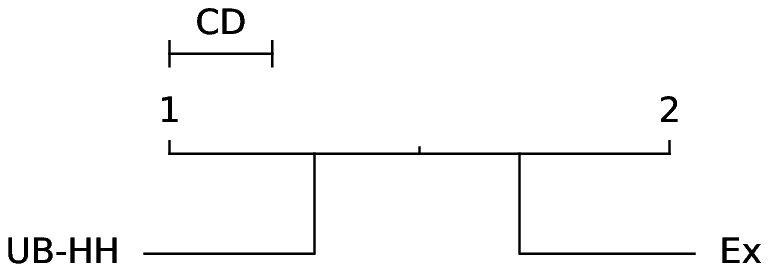}
         \caption{\textit{BS-Ex} and \textit{UB-HH}}
         \label{fig:bsex200}
     \end{subfigure}
    
     \begin{subfigure}[]{0.45\textwidth}
         \centering
         \includegraphics[width=\textwidth]{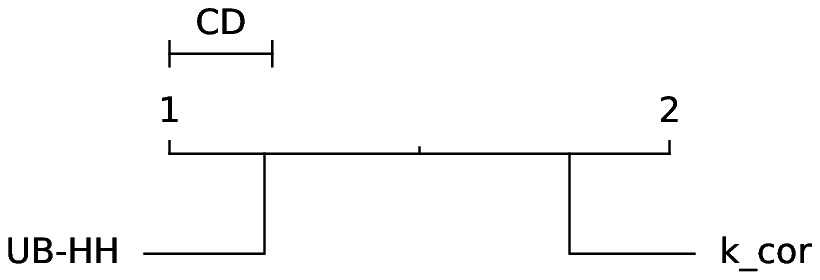}
         \caption{$k_{analytic}^{cor}$ and \textit{UB-HH}}
         \label{fig:kcor200}
     \end{subfigure}
     \begin{subfigure}[]{0.45\textwidth}
         \centering
         \includegraphics[width=\textwidth]{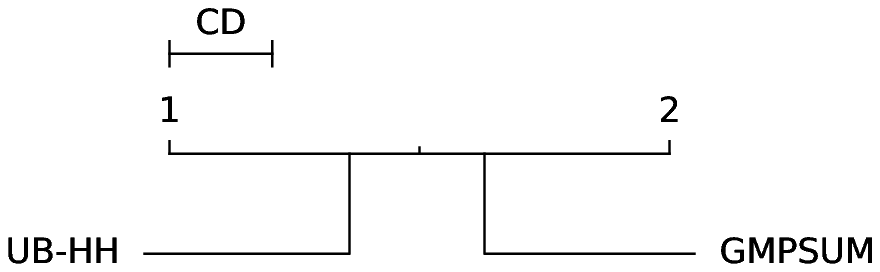}
         \caption{\textit{BS-GMPSUM} and \textit{UB-HH}}
         \label{fig:gmpsum200}
     \end{subfigure}
    
     \begin{subfigure}[]{0.45\textwidth}
         \centering
         \includegraphics[width=\textwidth]{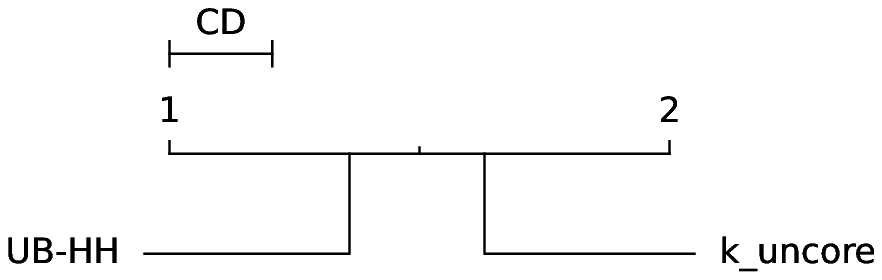}
         \caption{$k_{analytic}^{uncor}$ and \textit{UB-HH}}
         \label{fig:kuncor200}
     \end{subfigure}
     \begin{subfigure}[]{0.45\textwidth}
         \centering
         \includegraphics[width=\textwidth]{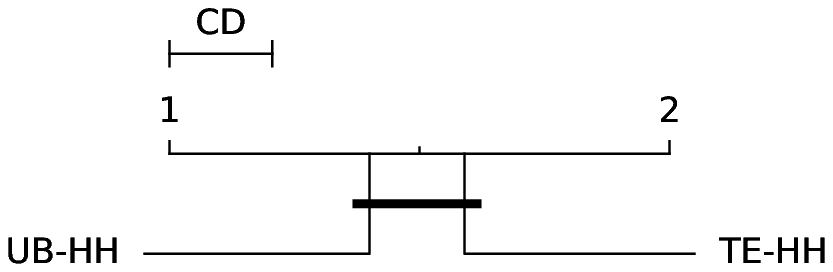}
         \caption{\textit{TE-HH} and \textit{UB-HH}}
         \label{fig:tehh200}
     \end{subfigure}
        \caption{Critical difference plot in \textit{balanced-time-quality} setting for all considered benchmarks}
        \label{fig:stat200}
\end{figure}

\textcolor{black}{In this section, we perform the statistical significance tests for all running scenarios: \textit{low-time}, \textit{high-quality}, and \textit{balanced-time-quality}. For this purpose, we use the Friedman~\cite{friedman40} test with the $5 \%$ error level. Also, the Nemenyi post-hoc~\cite{demvsar06} test is employed to check the pairwise significance of algorithms.} \newline
\textcolor{black}{Fig.~\ref{fig:stat200} shows the critical difference plot for all methods on all benchmarks on the \textit{balanced-time-quality} scenario. \textit{UB-HH} is significantly better than all methods except \textit{TE-HH}. In comparing \textit{UB-HH} and \textit{TE-HH}, these two algorithms perform equally on all benchmark datasets. However, in the case of uncorrelated benchmarks (i.e., \textit{ACO-Random}, \textit{ACO-Rat}, \textit{ACO-Virus}, \textit{ES}), the \textit{UB-HH} is significantly better than \textit{TE-HH}.
}

\begin{figure}[H]
     \centering
     \begin{subfigure}[]{0.45\textwidth}
         \centering
         \includegraphics[width=\textwidth]{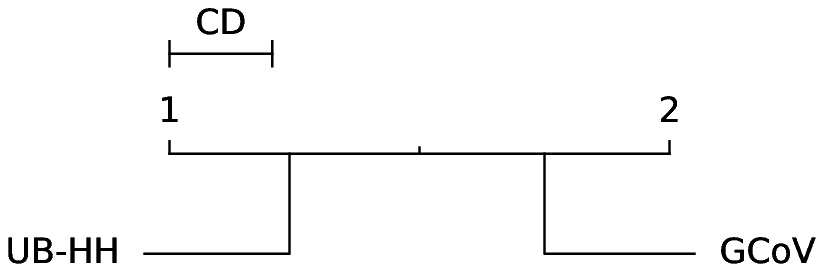}
         \caption{\textit{GCoV} and \textit{UB-HH}}
         \label{fig:gcov50}
     \end{subfigure}
     \begin{subfigure}[]{0.45\textwidth}
         \centering
         \includegraphics[width=\textwidth]{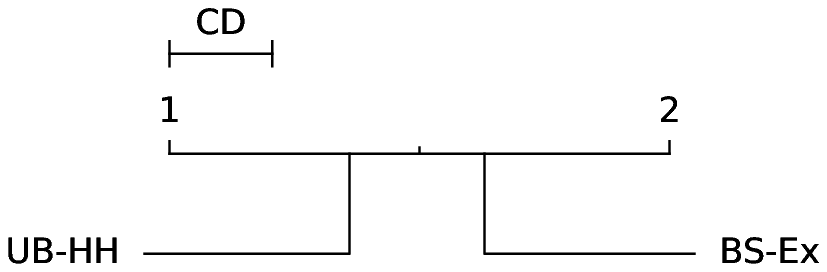}
         \caption{\textit{BS-Ex} and \textit{UB-HH}}
         \label{fig:bsex50}
     \end{subfigure}

     \begin{subfigure}[]{0.45\textwidth}
         \centering
         \includegraphics[width=\textwidth]{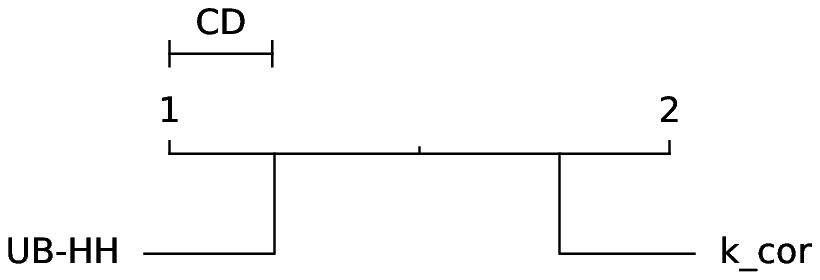}
         \caption{$k_{analytic}^{cor}$ and \textit{UB-HH}}
         \label{fig:kcor50}
     \end{subfigure}
     \begin{subfigure}[]{0.45\textwidth}
         \centering
         \includegraphics[width=\textwidth]{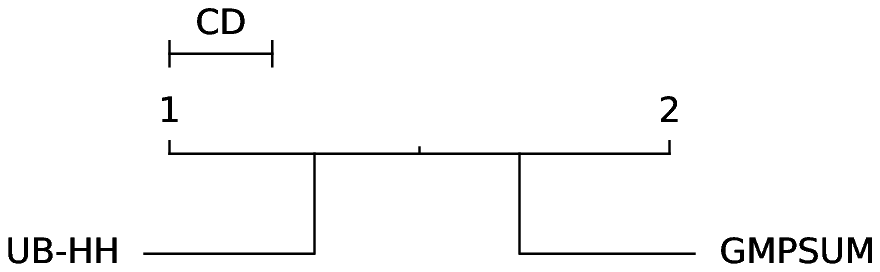}
         \caption{\textit{BS-GMPSUM} and \textit{UB-HH}}
         \label{fig:gmpsum50}
     \end{subfigure}

     \begin{subfigure}[]{0.45\textwidth}
         \centering
         \includegraphics[width=\textwidth]{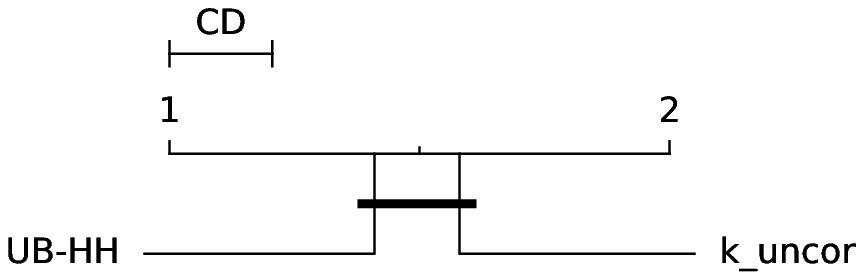}
         \caption{$k_{analytic}^{uncor}$ and \textit{UB-HH}}
         \label{fig:kuncor50}
     \end{subfigure}
        \caption{Critical difference plot in \textit{low-time} setting for all considered benchmarks}
        \label{fig:stat50}
\end{figure}

\textcolor{black}{Fig.~\ref{fig:stat50} shows the critical difference plot for all methods on all benchmarks in the \textit{low-time} scenario. \textit{UB-HH} is significantly better than other methods except for $k_{analytic}^{uncor}$, which they perform equally.
}

\begin{figure}[H]
     \centering
     \begin{subfigure}[]{0.45\textwidth}
         \centering
         \includegraphics[width=\textwidth]{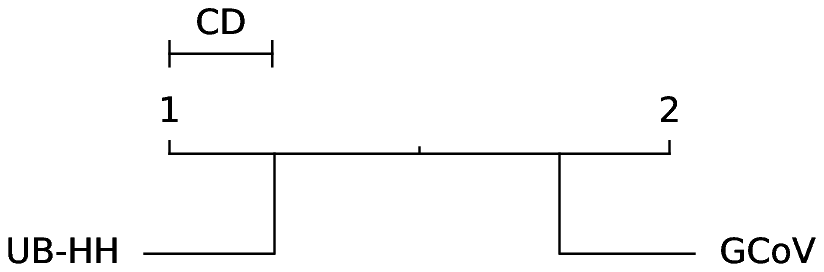}
         \caption{\textit{GCoV} and \textit{UB-HH}}
         \label{fig:gcov600}
     \end{subfigure}
     \begin{subfigure}[]{0.45\textwidth}
         \centering
         \includegraphics[width=\textwidth]{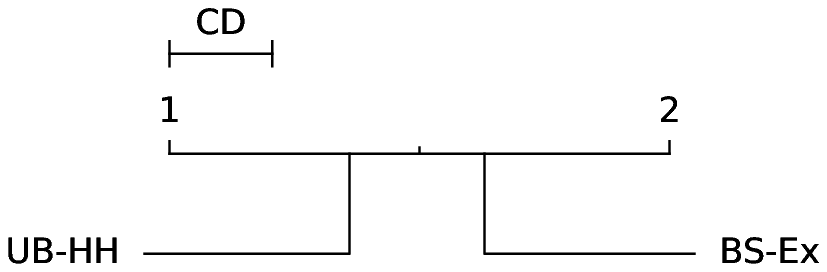}
         \caption{\textit{BS-Ex} and \textit{UB-HH}}
         \label{fig:bsex600}
     \end{subfigure}

     \begin{subfigure}[]{0.45\textwidth}
         \centering
         \includegraphics[width=\textwidth]{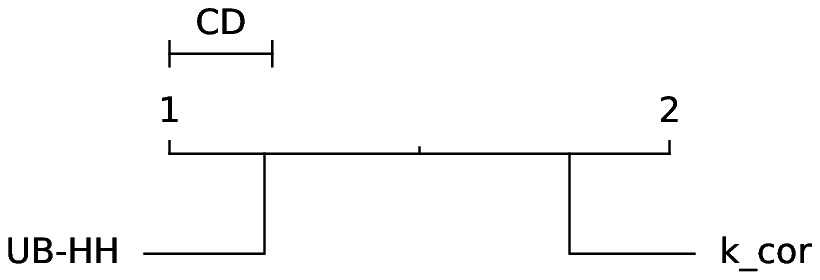}
         \caption{$k_{analytic}^{cor}$ and \textit{UB-HH}}
         \label{fig:kcor600}
     \end{subfigure}
     \begin{subfigure}[]{0.45\textwidth}
         \centering
         \includegraphics[width=\textwidth]{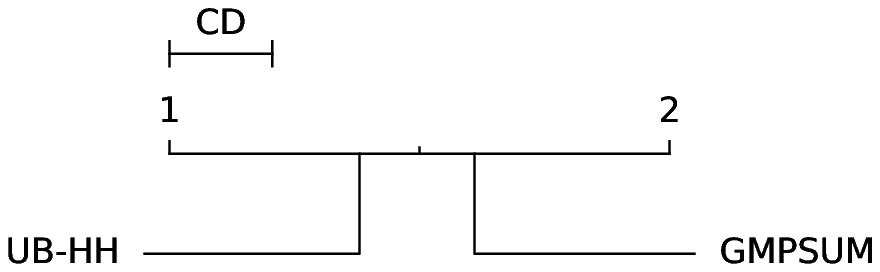}
         \caption{\textit{BS-GMPSUM} and \textit{UB-HH}}
         \label{fig:gmpsum600}
     \end{subfigure}

     \begin{subfigure}[]{0.45\textwidth}
         \centering
         \includegraphics[width=\textwidth]{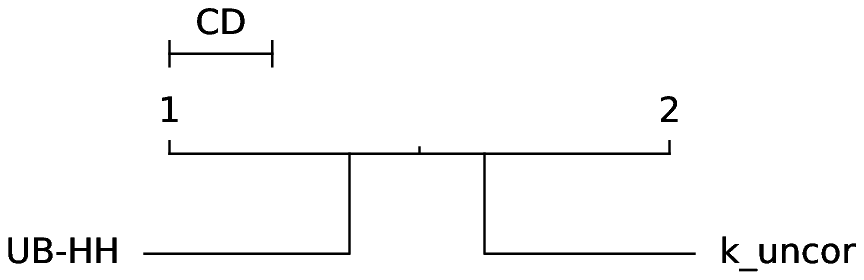}
         \caption{$k_{analytic}^{uncor}$ and \textit{UB-HH}}
         \label{fig:kuncor600}
     \end{subfigure}
        \caption{Critical difference plot in \textit{high-quality} setting for all considered benchmarks}
        \label{fig:stat600}
\end{figure}

\textcolor{black}{Fig.~\ref{fig:stat600} shows the critical difference plot for all methods on all benchmarks in the \textit{high-time} scenario. \textit{UB-HH} is significantly better than all methods.
}

\section{Conclusion}\label{conc}
In this work, we have proposed, the \textit{SCF} framework and $S^2D$ algorithm \textcolor{black}{to classify} a given set of strings for the first time. Set classification is an initial stage to choose the best matching LCS algorithm for a given set of strings. The $S^2D$ algorithm can classify all benchmarks in the \textit{LCS} literature with $98\%$ accuracy. By employing the $S^2D$ algorithm and the upper bound value, we have introduced the hyper-heuristic \textit{UB-HH} to identify the best heuristic for any set of strings. Unlike \textit{TE-HH}, the \textit{UB-HH} is very fast and independent of the ``search strategy''. The results show that our proposed hyper-heuristic (\textit{UB-HH}) outperforms \textit{TE-HH}\textcolor{black}{, even with the same base heuristics} in both solution quality and run time factors. In future work, it is possible to use other properties of strings, e.g., information measures or complexity standards, instead of upper bound to recognize the relevant heuristic. Furthermore, designing high-quality heuristics for correlated strings enriches the available methods of correlated datasets.

\section*{Acknowledgment}
We would like to show our gratitude to Bahare Adeli for helping design some of the figures.

\bibliography{mybibfile.bib}

\begin{thebibliography}{10}
\expandafter\ifx\csname url\endcsname\relax
  \def\url#1{\texttt{#1}}\fi
\expandafter\ifx\csname urlprefix\endcsname\relax\def\urlprefix{URL }\fi
\expandafter\ifx\csname href\endcsname\relax
  \def\href#1#2{#2} \def\path#1{#1}\fi

\bibitem{smith81}
T.~F. Smith, M.~S. Waterman, et~al., Identification of common molecular
  subsequences, Journal of molecular biology 147~(1) (1981) 195--197.

\bibitem{Jiang02}
T.~Jiang, G.~Lin, B.~Ma, K.~Zhang, A general edit distance between rna
  structures, Journal of computational biology 9~(2) (2002) 371--388.

\bibitem{shikder19}
R.~Shikder, P.~Thulasiraman, P.~Irani, P.~Hu, An openmp-based tool for finding
  longest common subsequence in bioinformatics, BMC research notes 12~(1)
  (2019) 1--6.

\bibitem{lu78}
S.-Y. Lu, K.~S. Fu, A sentence-to-sentence clustering procedure for pattern
  analysis, IEEE Transactions on Systems, Man, and Cybernetics 8~(5) (1978)
  381--389.

\bibitem{huang21}
J.~Huang, Z.~Fang, H.~Kasai, Lcs graph kernel based on wasserstein distance in
  longest common subsequence metric space, Signal Processing 189 (2021) 108281.

\bibitem{kruskal83}
J.~B. Kruskal, An overview of sequence comparison: Time warps, string edits,
  and macromolecules, SIAM review 25~(2) (1983) 201--237.

\bibitem{banerjee01}
A.~Banerjee, J.~Ghosh, Clickstream clustering using weighted longest common
  subsequences, in: Proceedings of the web mining workshop at the 1st SIAM
  conference on data mining, Vol. 143, Citeseer, 2001, p. 144.

\bibitem{dong18}
H.~Dong, J.~Man, L.~Jia, X.~Wang, Y.~Qin, K.~Liu, Traffic speed estimation
  using mobile phone location data based on longest common subsequence, in:
  2018 21st International Conference on Intelligent Transportation Systems
  (ITSC), IEEE, 2018, pp. 2819--2824.

\bibitem{Ding19}
J.~Ding, J.~Fang, Z.~Zhang, P.~Zhao, J.~Xu, L.~Zhao, Real-time trajectory
  similarity processing using longest common subsequence, in: 2019 IEEE 21st
  International Conference on High Performance Computing and Communications;
  IEEE 17th International Conference on Smart City; IEEE 5th International
  Conference on Data Science and Systems (HPCC/SmartCity/DSS), 2019, pp.
  1398--1405.
\newblock \href {http://dx.doi.org/10.1109/HPCC/SmartCity/DSS.2019.00194}
  {\path{doi:10.1109/HPCC/SmartCity/DSS.2019.00194}}.

\bibitem{maier78}
D.~Maier, The complexity of some problems on subsequences and supersequences,
  Journal of the ACM (JACM) 25~(2) (1978) 322--336.

\bibitem{djukanovic19}
M.~Djukanovic, G.~R. Raidl, C.~Blum, A beam search for the longest common
  subsequence problem guided by a novel approximate expected length
  calculation, in: International Conference on Machine Learning, Optimization,
  and Data Science, Springer, 2019, pp. 154--167.

\bibitem{Abdi2022}
A.~Abdi, M.~Hooshmand, Longest common subsequence: Tabular vs. closed-form
  equation computation of subsequence probability, arXiv preprint
  arXiv:2206.11726.

\bibitem{blum07}
C.~Blum, M.~J. Blesa, Probabilistic beam search for the longest common
  subsequence problem, in: International Workshop on Engineering Stochastic
  Local Search Algorithms, Springer, 2007, pp. 150--161.

\bibitem{norvig92}
P.~Norvig, Paradigms of artificial intelligence programming: case studies in
  Common LISP, Morgan Kaufmann, 1992.

\bibitem{blum09}
C.~Blum, M.~J. Blesa, M.~L{\'o}pez-Ib{\'a}{\~n}ez, Beam search for the longest
  common subsequence problem, Computers \& Operations Research 36~(12) (2009)
  3178--3186.

\bibitem{mousavi12improved}
S.~R. Mousavi, F.~Tabataba, An improved algorithm for the longest common
  subsequence problem, Computers \& Operations Research 39~(3) (2012) 512--520.

\bibitem{dragon}
A.~Aho, M.~Lam, R.~Sethi, J.~Ullman,
  \href{https://books.google.nl/books?id=dIU\_AQAAIAAJ}{Compilers: Principles,
  Techniques, \& Tools}, Pearson/Addison Wesley, 2007.
\newline\urlprefix\url{https://books.google.nl/books?id=dIU\_AQAAIAAJ}

\bibitem{Nikolic21}
B.~Nikolic, A.~Kartelj, M.~Djukanovic, M.~Grbic, C.~Blum, G.~Raidl,
  \href{https://www.mdpi.com/2227-7390/9/13/1515}{Solving the longest common
  subsequence problem concerning non-uniform distributions of letters in input
  strings}, Mathematics 9~(13).
\newblock \href {http://dx.doi.org/10.3390/math9131515}
  {\path{doi:10.3390/math9131515}}.
\newline\urlprefix\url{https://www.mdpi.com/2227-7390/9/13/1515}

\bibitem{burke2003}
E.~Burke, G.~Kendall, J.~Newall, E.~Hart, P.~Ross, S.~Schulenburg,
  Hyper-heuristics: An emerging direction in modern search technology, in:
  Handbook of metaheuristics, Springer, 2003, pp. 457--474.

\bibitem{tabataba12hyper}
F.~S. Tabataba, S.~R. Mousavi, A hyper-heuristic for the longest common
  subsequence problem, Computational biology and chemistry 36 (2012) 42--54.

\bibitem{cov98}
B.~Everitt, The cambridge dictionary of statistics cambridge university press,
  Cambridge, UK Google Scholar.

\bibitem{gusfield97}
D.~Gusfield, Algorithms on stings, trees, and sequences: Computer science and
  computational biology, Acm Sigact News 28~(4) (1997) 41--60.

\bibitem{shyu09}
S.~J. Shyu, C.-Y. Tsai, Finding the longest common subsequence for multiple
  biological sequences by ant colony optimization, Computers \& Operations
  Research 36~(1) (2009) 73--91.

\bibitem{Easton08}
T.~Easton, A.~Singireddy, A large neighborhood search heuristic for the longest
  common subsequence problem, J. Heuristics 14 (2008) 271--283.
\newblock \href {http://dx.doi.org/10.1007/s10732-007-9038-y}
  {\path{doi:10.1007/s10732-007-9038-y}}.

\bibitem{fayed19}
H.~A. Fayed, A.~F. Atiya, Speed up grid-search for parameter selection of
  support vector machines, Applied Soft Computing 80 (2019) 202--210.

\bibitem{friedman40}
M.~Friedman, A comparison of alternative tests of significance for the problem
  of m rankings, The Annals of Mathematical Statistics 11~(1) (1940) 86--92.

\bibitem{demvsar06}
J.~Dem{\v{s}}ar, Statistical comparisons of classifiers over multiple data
  sets, The Journal of Machine learning research 7 (2006) 1--30.

\end{thebibliography}

\end{document}